\documentclass{article}

\usepackage[
    final,
    nonatbib,
]{neurips_2024}

\usepackage{float}
\usepackage{wrapfig}
\usepackage{graphicx}
\graphicspath{{./figs}}

\usepackage{amsmath, amssymb, amsfonts, mathtools}
\usepackage{bm}

\DeclareMathOperator*{\argmin}{argmin_{\mathnormal{}}}
\DeclareMathOperator{\EE}{\mathbb{E}}
\DeclareMathOperator{\RR}{\mathbb{R}}
\DeclareMathOperator{\ZZ}{\mathbb{Z}}

\DeclareMathOperator{\CC}{\mathcal{C}}

\DeclareMathOperator{\GG}{\mathcal{G}}

\DeclareMathOperator{\LL}{\mathcal{L}}

\DeclareMathOperator{\NN}{\mathcal{N}}
\DeclareMathOperator{\PP}{\mathcal{P}}

\newcommand{\norm}[1]{\left\lVert#1\right\rVert}
\newcommand{\kl}[2]{\mathcal{D}_{\mathtt{KL}}\left({#1}\,\|\,{#2}\right)}
\newcommand{\KL}[2]{\mathcal{D}_{\mathtt{KL}}\Big{(}{#1}\,\big\|\,{#2}\Big{)}}
\newcommand{\expec}[2]{\EE_{#1} [ {#2} ]}
\newcommand{\expect}[2]{\EE_{#1}\!\Big{[} {#2} \Big{]}}
\newcommand{\EXPECT}[2]{\EE_{#1}\!\Bigg{[} {#2} \Bigg{]}}

\def\eqref#1{equation~\ref{#1}}

\def\1{\bm{1}}

\def\eps{{\epsilon}}

\def\rr{{\textnormal{r}}}

\DeclareMathAlphabet{\mathsfit}{\encodingdefault}{\sfdefault}{m}{sl}
\SetMathAlphabet{\mathsfit}{bold}{\encodingdefault}{\sfdefault}{bx}{n}

\def\entry#1#2{${#1}\scriptstyle{\pm{#2}}$}   

\usepackage{xcolor}
\definecolor{gr}{HTML}{37C532}
\definecolor{grey}{HTML}{4D4D4D}
\definecolor{gray}{HTML}{4D4D4D}
\definecolor{color_enc}{HTML}{ED1C24}
\definecolor{color_dec}{HTML}{29ABE2}
\definecolor{color_decenc}{HTML}{dd1aff}
\def\enc#1{{\color{color_enc}{#1}}\xspace}
\def\dec#1{{\color{color_dec}{#1}}\xspace}
\def\decenc#1{{\color{color_decenc}{#1}}\xspace}

\usepackage{xspace}
\newcommand{\cvae}{$\CC$-VAE\xspace}
\newcommand{\lvae}{$\LL$-VAE\xspace}
\newcommand{\gvae}{$\GG$-VAE\xspace}
\newcommand{\grelu}{\gvae$_\mathrm{+relu}$\xspace}
\newcommand{\gexp}{\gvae$_\mathrm{+exp}$\xspace}
\newcommand{\cifar}{CIFAR$_{16\times16}$\xspace}

\DeclareRobustCommand{\xx}{\bm{x}}
\DeclareRobustCommand{\zz}{\bm{z}}
\DeclareRobustCommand{\rr}{\bm{r}}

\DeclareRobustCommand{\bmu}{\bm{\mu}}
\DeclareRobustCommand{\bsig}{\bm{\sigma}}

\DeclareRobustCommand{\dr}{\bm{\delta r}}

\newcommand{\pois}{\mathcal{P}\mathrm{ois}}
\newcommand{\pvae}{$\PP$-VAE\xspace}

\def\architecture#1#2{$\left< {\color{color_enc}\mathtt{{#1}}} \vert {\color{color_dec}\mathtt{{#2}}} \right>$}

\newcommand{\linlin}{\architecture{lin}{lin}\xspace}

\newcommand{\convconv}{\architecture{conv}{conv}\xspace}
\newcommand{\convlin}{\architecture{conv}{lin}\xspace}

\def\entry#1#2{${#1}\scriptstyle{\pm{#2}}$}

\usepackage{hyperref}
\usepackage{lipsum}
\usepackage{url}             
\usepackage{booktabs}        
\usepackage{amsfonts}        
\usepackage{nicefrac}        
\usepackage{microtype}       
\usepackage{pifont}          
\usepackage{siunitx}         
\usepackage[normalem]{ulem}  
\usepackage{extarrows}

\usepackage{dirtytalk}
\usepackage{wrapfig}
\usepackage{tikz}
\usepackage{caption}
\usepackage{bbding}   
\usepackage{enumitem} 
\usepackage{empheq}   

\usepackage{algorithm}
\usepackage{algpseudocode}

\usepackage{tabularray}
\usepackage{tabularx}
\usepackage{array}  
\newcolumntype{C}{>{\centering\arraybackslash}X}  

\usepackage{cleveref}
\crefname{figure}{Fig.}{Figs.}
\Crefname{figure}{Figure}{Figures}

\usepackage[utf8]{inputenc} 
\usepackage[
    backend=biber,
    sorting=none,
    style=numeric-comp,
    maxnames=2,
    minnames=1,
]{biblatex}
\usepackage[T1]{fontenc}    

\usetikzlibrary{shapes}
\usepackage{silence}
\title{Poisson Variational Autoencoder}

\author{
    Hadi Vafaii$^1$ \hspace{21mm} Dekel Galor$^1$ \hspace{19mm} Jacob L.~Yates$^1$ \\
    \hspace{-0.45cm} \href{mailto:vafaii@berkeley.edu}{\texttt{vafaii@berkeley.edu}} \hspace{7mm} \texttt{galor@berkeley.edu} \hspace{7mm} \texttt{yates@berkeley.edu} \\[1.0em]
    $^1$UC Berkeley\\
}
\begin{document}

\vspace*{-10mm}
\maketitle

\begin{abstract}
Variational autoencoders (VAEs) employ Bayesian inference to interpret sensory inputs, mirroring processes that occur in primate vision across both ventral \cite{higgins2021unsupervised} and dorsal \cite{vafaii2023hierarchical} pathways. Despite their success, traditional VAEs rely on continuous latent variables, which deviates sharply from the discrete nature of biological neurons. Here, we developed the Poisson VAE (\pvae), a novel architecture that combines principles of predictive coding with a VAE that encodes inputs into discrete spike counts. Combining Poisson-distributed latent variables with predictive coding introduces a metabolic cost term in the model loss function, suggesting a relationship with sparse coding which we verify empirically. Additionally, we analyze the geometry of learned representations, contrasting the \pvae to alternative VAE models. We find that the \pvae encodes its inputs in relatively higher dimensions, facilitating linear separability of categories in a downstream classification task with a much better ($5\times$) sample efficiency. Our work provides an interpretable computational framework to study brain-like sensory processing and paves the way for a deeper understanding of perception as an inferential process.
\end{abstract}

\begin{figure}[h!]
    \centering
    \includegraphics[width=\linewidth]{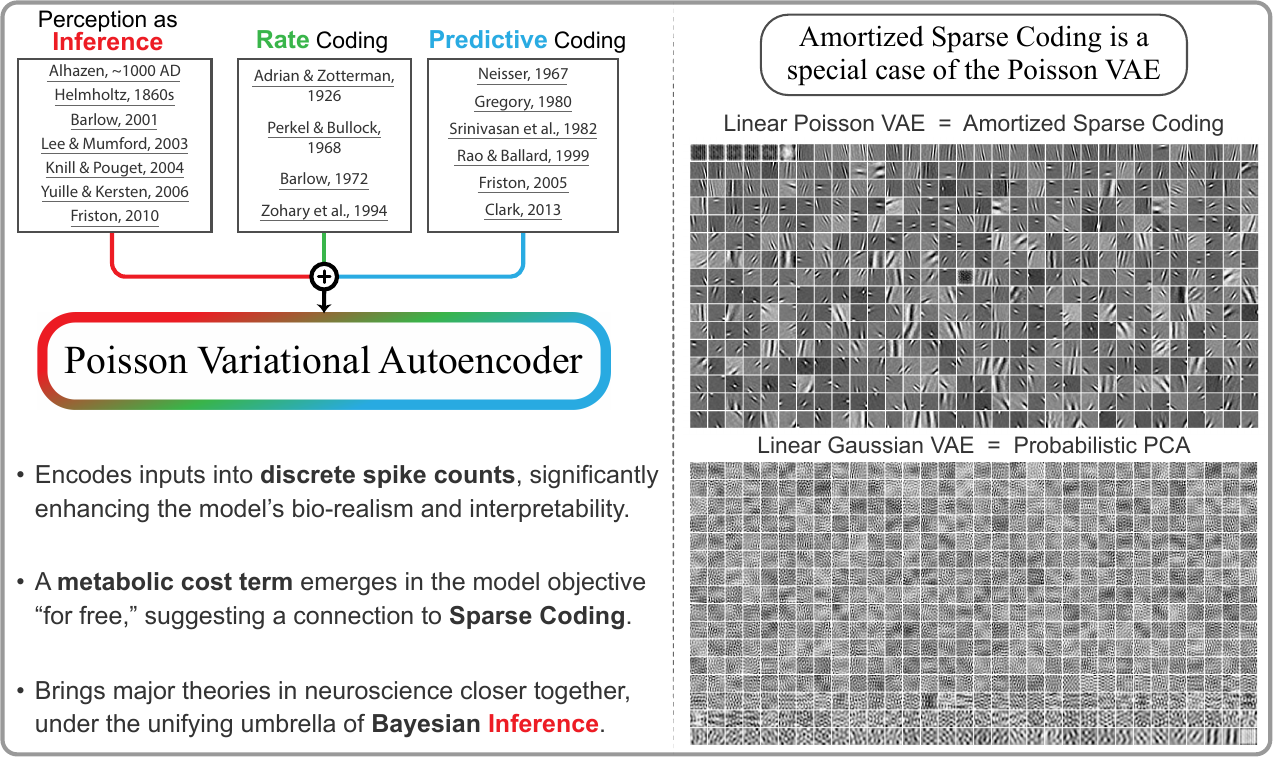}
    \caption{
        Graphical abstract. Introducing the Poisson Variational Autoencoder (\pvae), which draws on key concepts in neuroscience.
        When trained on natural image patches, \pvae with a linear decoder develops Gabor-like feature selectivity, reminiscent of Sparse Coding \cite{olshausen1996emergence}. In sharp contrast, the standard Gaussian VAE learns the principal components \cite{tipping1999prob}. Our code, data, and model checkpoints are available at this repository: \href{https://github.com/hadivafaii/PoissonVAE}{\color{purple}https://github.com/hadivafaii/PoissonVAE}
    }
    \label{fig:grphical-abstract}
\end{figure}

\section{Introduction}

The study of artificial neural networks (ANN) and neuroscience has always been closely linked, driving advancements in both fields \cite{mcculloch1943logical, churchland1988perspectives, thomas2008connectionist, kriegeskorte2015deep, lindsay2021convolutional, marino2022predictive}. Despite the close proximity of the two fields, most ANN models deviate substantially from biological brains \cite{bowers2023deep, wichmann2023are}. A major challenge is designing models that not only perform well computationally but also exhibit \say{brain-like} structure and function. This is seen both as a goal for improving ANNs \cite{zador2023catalyzing, sinz2019engineering, hassabis2017neuroscience}, and better understanding biological brains \cite{lindsay2021convolutional, kriegeskorte2015deep, kanwisher2023using, richards2022application, richards2019deep, barrett2019analyzing}, which has recently been referred to as the \textit{neuroconnectionist} research programme \cite{doerig2023neuroconnectionist}.

Drawing from neuroscience, a major guiding idea is that perception is a process of inference \cite{alhazen, helmholtz1867handbuch}, where the brain constructs a representation of the external world by inferring the causes of sensory inputs \cite{lee2003hierarchical, olshausen2014, boring1946perception, friston2010free}. This concept is mirrored in \say{generative AI} where models learn the generative process underlying their inputs \cite{bond2021deep, chan2024tutorial, zhao2023survey}. However, in this vein, there is a tension between small well-understood models that are directly inspired by cortex, such as sparse coding \cite{olshausen1996emergence} and predictive coding \cite{rao1999predictive}, and deep generative models that perform well \cite{rombach2022high, karras2019style, vahdat2020nvae, child2021vdvae}.
 
The variational autoencoder (VAE; \cite{kingma2014auto, rezende2014stochastic}) model family is a promising candidate for neuroconnectionist goals for multiple reasons. First, VAEs learn probabilistic generative models of their inputs and are grounded in Bayesian probability theory, providing a solid theoretical foundation that directly incorporates the concept of perceptual inference \cite{marino2022predictive, helmholtz1867handbuch}. Second, the VAE model family, specifically hierarchical VAEs, is broad with other generative models, such as diffusion models, understood as special cases of hierarchical VAEs \cite{kingma2023understanding, ldm_tutorial_2023, kingma2021variational}. Finally, VAEs learn representations that are similar to cortex \cite{higgins2021unsupervised, vafaii2023hierarchical, csikor2023topdown}, exhibit cortex-like topographic organization \cite{keller2021modeling, keller2021topographic}, and make perceptual errors that mimic those of humans \cite{storrs2021unsupervised}, indicating a significant degree of neural, organizational, and psychophysical alignment with the brain.

However, standard VAEs diverge from brains in the way they encode information. Biological neurons fire all-or-none action potentials \cite{adrian1932nobel}, and are thought to represent information via firing rate \cite{Adrian1926TheIP, perkel1968neural, barlow1972single, zohary1994correlated, rieke1999spikes}. These firing rates must be positive and generate discrete \say{spike} counts, which exhibit conditionally Poisson-like statistics in small counting windows \cite{teich1989fractal, rieke1999spikes, goris2014partitioning}. In contrast, VAEs are typically parameterized with real-valued, continuous, Gaussian distributions \cite{kingma2019introduction}.

\begin{wrapfigure}[28]{r}{0.60\textwidth}
    \centering
    \includegraphics[width=0.60\textwidth]{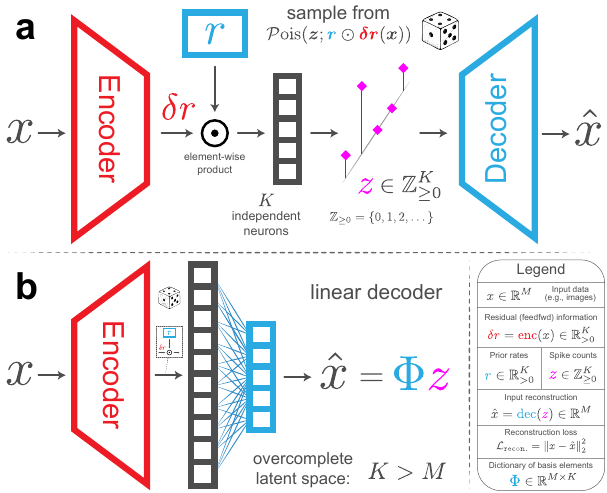}
    \caption{
        \textbf{(a)} Model architecture. Colored shapes indicate learnable model parameters, including the prior firing rates, $\dec{\rr}$. We color code the model's \enc{inference} and \dec{generative} components using \enc{red} and \dec{blue}, respectively. The \pvae encodes its inputs in discrete spike counts, $\zz$, significantly enhancing its biological realism. \textbf{(b)} ``Amortized Sparse Coding'' is a special case within the \pvae model family: it's a \pvae with a linear decoder and an overcomplete latent space.
    }
    \label{fig:archi+amort}
\end{wrapfigure}

\paragraph{Contributions.} In this work, we address this discrepancy by introducing the Poisson Variational Autoencoder (\pvae), a novel architecture that combines perceptual inference with two other inspirations from neuroscience (\cref{fig:grphical-abstract}). First, that information is encoded in the rates of discrete spike counts, which are approximately Poisson-distributed on short time intervals. And second, that feedforward connections encode deviations from expectations contained in feedback connections (\cref{fig:archi+amort}a; \cite{rao1999predictive, gilbert2013top}). We introduce a reparameterization trick for Poisson samples (\Cref{algo:rsample}), and derive the evidence lower bound (ELBO) objective for the \pvae (\cref{eq:pvae-nelbo}).  Overall, we believe \pvae introduces a promising new model at the intersection of computational neuroscience and machine learning that offers several appealing features over existing VAE architectures:

\begin{itemize}
    \item The \pvae loss derivation (\cref{eq:pvae-nelbo}) naturally results in a metabolic cost term that penalizes high firing rates, such that \pvae with a linear decoder implements amortized sparse coding (\cref{fig:archi+amort}b). We validate this prediction empirically.
    \item \pvae largely avoids the prevalent posterior collapse issue, maintaining many more active latents compared to alternative VAE models (\Cref{tab:models}), especially the continuous ones.
    \item \pvae encodes its inputs in relatively higher dimensions, facilitating linear separability of categories in a downstream classification task with a much better ($5\times$) sample efficiency.
\end{itemize}

We evaluate these results on two natural image datasets and MNIST. The \pvae paves the way for the future development of interpretable hierarchical models that perform \say{brain-like} inference.

\section{Background \& Related work}

\paragraph{Perception as inference: connections to neuroscience and machine learning.}  A centuries-old idea \cite{alhazen, helmholtz1867handbuch}, \say{perception as inference} argues that coherent perception of the world results from the unconscious inference over the causes of the senses. In other words, the brain learns a generative model of the sensory inputs. This has led to fruitful theoretical work in neuroscience \cite{lee2003hierarchical, knill2004bayesian, barlow2001redundancy, friston2009free} and machine learning \cite{dayan1995helmholtz, lotter2017deep}, including VAEs \cite{kingma2019introduction}. See \textcite{marino2022predictive} for a review.

\paragraph{Efficient, predictive, and sparse coding.} 
Another longstanding idea in neuroscience is that brains are adapted to the statistics of the environment. Efficient coding states that brains represent as much information about the environment as possible while minimizing neural resource use \cite{attneave1954some, barlow1961possible}. 

Predictive coding \cite{srinivasan1982predictive, rao1999predictive, friston2005theory} postulates that the brain generates a statistical prediction of its inputs, with feedforward networks carrying only the prediction errors or unexplained information \cite{clark2013whatever}. More recently, ANNs based on predictive coding have been shown to capture a wide range of phenomena in biological neurons across the visual system \cite{lotter2020neural, millidge2024predictive}. More broadly, prediction in time has emerged as an objective that lends itself to brain-like representations \cite{singer2023hierarchical, fiquet2024polar}.

Sparse coding (SC) is directly inspired by efficient coding, aiming to explain inputs as sparsely as possible \cite{barlow1972single, olshausen2004sparse}. SC was the first unsupervised model to learn representations closely resembling the receptive fields of V1 neurons \cite{olshausen1996emergence} and predicts an array of empirical features of neural activity \cite{vinje2000sparse, barth2012experimental, quiroga2008sparse, hromadka2008sparse, poo2009odor, wolfe2010sparse, willmore2011sparse, haider2010synaptic, crochet2011synaptic, petersen2019sensorimotor, froudarakis2014population}. SC is formalized with a generative model where neural activations $\zz$ are sampled from a sparsity-inducing prior, $\zz \sim p(\zz)$, and the input image $\xx$ is reconstructed as a linear combination of basis vectors $\bm{\Phi}$, plus additive Gaussian noise, $\hat{\xx} = \bm{\Phi} \zz + \bm{\varepsilon}$. The SC loss is as follows:

\begin{equation}\label{eq:sparse-coding}
    \LL_\mathrm{SparseCoding}\left(
        \xx;
        \bm{\Phi},
        \zz
    \right)
    =
    \norm{\xx - \bm{\Phi} \zz}_2^2
    +
    \beta \norm{\zz}_1.
\end{equation}


Commonly used algorithms for sparse coding include the locally competitive algorithm (LCA; \cite{rozell2008sparse}), which is a biologically plausible algorithm to optimize \cref{eq:sparse-coding}, and iterative shrinkage-thresholding algorithm (ISTA; \cite{daubechies2004iterative, fista2009}), which has shown robust performance in learning sparse codes given a fixed dictionary $\bm{\Phi}$.

\paragraph{VAE objective.} VAEs define a probabilistic generative model $p(\xx, \zz)$, where $\xx$ denotes the observed data and $\zz$ are some latent variables. The generative process samples $\zz$ from a prior distribution $p(\zz)$ and then generates the observed data $\xx$ from the conditional distribution $p_{\bm{\theta}}(\xx|\zz)$, also known as the \say{decoder}. The \say{encoder}, $q_{\bm{\phi}}(\zz|\xx)$, performs approximate inference on the inputs. Model parameters are learned by maximizing the evidence lower bound (ELBO) objective, which is derived from variational inference (see \cref{sec:supp-derivations} for the full set of derivations). The ELBO is given by:

\begin{equation}\label{eq:elbo}
 \log p(\xx) \geq \expect{q_{\bm{\phi}}(\zz \vert \xx)} {\log p_{\bm{\theta}}(\xx \vert \zz)} - \KL{q_{\bm{\phi}}(\zz \vert \xx)}{p(\zz)} = \LL_\text{VAE}(\xx; \bm{\theta}, \bm{\phi}).
\end{equation}

The first term captures the reconstruction performance of the decoder, and the second term, the \say{$\mathtt{KL}$ term,} captures the divergence of the approximate posterior from the prior.

The specific form of these distributions is up to the practitioner. In standard VAEs, factorized Gaussians are typically used: $q = \NN(\zz; \bmu(\xx), \bsig^2(\xx))$ and $p = \NN(\zz; \bm{0}, \bm{1})$. The likelihood, $p_{\bm{\theta}}(\xx|\zz)$, is also typically modeled as a Gaussian conditioned on a parameterized neural network $\mathrm{dec}_{\bm{\theta}}(\zz)$.

\paragraph{Amortized inference in VAEs.} A major contribution of VAEs is the idea of amortizing inference over the latents $\zz$ with a black box ANN \cite{ganguly2023amortized, amos2023tutorial}. \say{Amortized} inference borrows a term from finance to capture the idea of spreading out costs---here, the cost of performing inference over multiple samples. In amortized inference, a neural network learns (during training) how to map a data sample to a distribution over latent variables given the sample. The cost is paid during training, but the trained model can then be used to perform inference on future samples efficiently. It has been argued that the brain performs amortized inference for computational efficiency \cite{gershman2014amortized}. 

\paragraph{VAEs connection to biology.} VAEs have been shown to contain individual latents that resemble neurons, capturing a wide range of the phenomena observed in visual cortical areas \cite{csikor2023topdown} and human perceptual judgments \cite{storrs2021unsupervised}. Like many other ANN models \cite{conwell2023can, elmoznino2022high}, VAEs have been found to learn representations that are predictive of single-neuron activity in both the ventral \cite{higgins2021unsupervised} and dorsal \cite{vafaii2023hierarchical} streams. However, unlike most ANNs, the mapping from certain VAEs to neural activity is incredibly sparse, even one-to-one in some cases \cite{higgins2021unsupervised, vafaii2023hierarchical}.

\paragraph{Discrete VAEs.} VAEs with discrete latent spaces, such as VQ-VAE \cite{van2017neural} and Categorical VAE \cite{jang2017categorical}, are designed to capture complex data structures by mapping inputs to a finite set of latent variables. Unlike traditional VAEs that use continuous latent spaces, these models leverage discrete representations to enhance interpretability and can yield high performance with lower capacity \cite{kamata2022fully}.

\paragraph{VAEs connection to sparse coding.} Previous work has attempted to connect sparse coding and VAEs directly \cite{geadah2024sparse, tonolini2020variational, xiao2024scvae}, with each approaching the problem differently. \textcite{geadah2024sparse} introduced sparsity-inducing priors (such as Laplace or Cauchy) and a linear decoder with an overcomplete latent space. \textcite{tonolini2020variational} introduced a spike and slab prior into a modified ELBO, and \textcite{xiao2024scvae} added a sparse coding layer learned by ISTA to the latent space of a VQ-VAE. Notably, none of the three ended up minimizing the sparse coding loss. Two of the three maintain the linear generative model with an overcomplete latent space, but the ELBO in both requires an additional approximation step for the $\mathtt{KL}$ term \cite{geadah2024sparse, tonolini2020variational}.

\section{Introducing the Poisson Variational Autoencoder (\texorpdfstring{\pvae}{P-VAE})}\label{sec:theory}
Our main contribution is integrating Poisson-distributed latents into VAEs, where both the approximate posterior and the prior are parameterized as Poisson distributions. Critically, the latents $\zz$ are no longer continuous variables, but rather they are discrete spike counts. To perform inference over discrete latents, we introduce a Poisson reparameterization trick. We then derive the $\mathtt{KL}$ term and obtain the full \pvae objective.

\paragraph{Poisson reparameterization trick.} For a homogeneous Poisson process \cite{cox1980point, shchur2020fast, shchur2022modeling}, given a window size $\Delta t = 1$, and rate $\lambda$, we can generate Poisson distributed counts by drawing randomly distributed wait-times from an exponential distribution with mean $1 / \lambda$ and counting all events where the cumulative time is less than 1. Because the exponential distribution is trivially reparameterized \cite{kingma2014auto}, and PyTorch contains an implementation \cite{paszke2019pytorch}, we need only to approximate the hard threshold for comparing cumulative wait times with the window size. We accomplish this by replacing the indicator function with a sigmoid as in refs.~\cite{jang2017categorical, maddison2017concrete}.

\begin{algorithm}[t!]
\caption{Reparameterized sampling (rsample) for Poisson distribution.}
\label{algo:rsample}
\begin{algorithmic}[1]
\Statex \textbf{Input:}
\Statex $\bm{\lambda} \in \mathbb{R}_{>0}^{B \times K}$ \hspace{8.8mm}
{\color{gray}\# rate parameter; $B$, batch size; $K$, latent dimensionality}
\Statex $n\_\mathrm{exp}$ \hspace{15.7mm}
{\color{gray}\# number of exponential samples to generate}
\Statex $\mathrm{temperature}$ \hspace{6mm}
{\color{gray}\# controls the sharpness of the thresholding}
\vspace{1mm}
\Procedure{Rsample}{$\bm{\lambda}, n\_\mathrm{exp}, \mathrm{temperature}$}
    \State $\mathrm{Exp} \gets \mathrm{Exponential}(\bm{\lambda})$
    \Comment{{\color{gray}create exponential distribution}}
    \State $\Delta t \gets \mathrm{Exp}.\text{rsample}((n\_\mathrm{exp},))$
    \Comment{{\color{gray}sample inter-event times, $\Delta t: \left[ n\_\mathrm{exp} \times B \times K\right ]$}}
    \State $\mathrm{times} \gets \text{cumsum}(\Delta t, \text{dim=0})$ \Comment{{\color{gray}compute arrival times, same shape as $\Delta t$}} 
    \State $\mathrm{indicator} \gets \text{sigmoid}\left(\frac{1 - \mathrm{times}}{\mathrm{temperature}}\right)$
    \Comment{{\color{gray}soft indicator for events within unit time}}
    \State $\zz \gets \text{sum}(\mathrm{indicator}, \text{dim=0})$ \Comment{{\color{gray}event counts, or number of spikes, $\zz: \left[ B \times K\right ]$}}
    \State \textbf{return} $\zz$
    %
\EndProcedure
\end{algorithmic}
\end{algorithm}


\begin{wrapfigure}[16]{r}{0.4\textwidth}
    \begin{center}
        \includegraphics[width=0.4\textwidth]{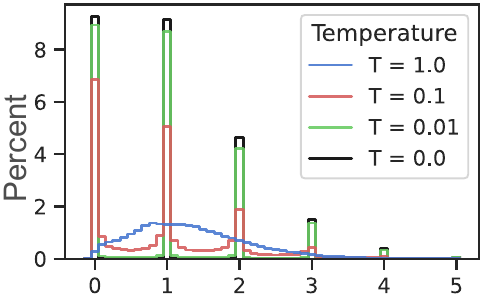}
    \end{center}
    \caption{
        Relaxed Poisson distribution. Samples are drawn using \Cref{algo:rsample} for $\lambda = 1$. At non-zero temperatures, samples are non-integer, but approach the true Poisson distribution as $T \rightarrow 0$.
    }
    \label{fig:relaxed_poisson}
\end{wrapfigure}

\Cref{algo:rsample} demonstrates the steps: Given a matrix of rates $\bm{\lambda}$, sample $n\_\mathrm{exp}$ wait times $t_1, t_2, ... t_{n\_\mathrm{exp}}$ for each element of $\bm{\lambda}$ by sampling from an exponential distribution with mean $1 / \bm{\lambda}$. We then calculate the cumulative event times $S(n\_\mathrm{exp}) = \sum_{j=1}^{n\_\mathrm{exp}} t_j$, pass them through a sigmoid $\sigma(\frac{1-S}{\mathrm{temperature}})$, and sum over samples to get event counts, $\zz$. The temperature controls the sharpness of the thresholding. We adaptively scale the number of samples, $n\_\mathrm{exp}$, by keeping track of the maximum rate in each batch, $\lambda_\text{max}$, and then use the inverse cumulative density function ($\mathrm{cdf}$) for Poisson to find the number of samples, $n\_\mathrm{exp}$, such that $\mathrm{cdf}(n\_\mathrm{exp};\lambda_\textrm{max}) = 0.99999$.

At non-zero temperatures, our parameterization algorithm provides a continuous relaxation of the Poisson distribution. \Cref{fig:relaxed_poisson} shows histograms of samples drawn using \Cref{algo:rsample} for rate $\lambda = 1$ and temperatures $T = 1.0, 0.1, 0.01$, and $0$. The latter case ($T = 0$, true Poisson) is equivalent to \texttt{torch.poisson()}.


\paragraph{\pvae architecture and residual parameterization.} The architecture of \pvae captures the interactions between feedforward and feedback connections that are present in all visual cortical areas \cite{felleman1991distributed, markov2014anatomy}. Feedforward areas carry sensory information and feedback connections are thought to carry modulatory signals such as attention \cite{gilbert2013top} or prediction \cite{rao1999predictive}, which interact multiplicatively with feedforward inputs \cite{gilbert2013top, disney2021neuromodulatory}.

\pvae embodies this idea by having the posterior rates depend on the prior, such that $\rr_\text{prior} = \dec{\rr}$ and $\rr_\text{post.} = \dec{\rr} \odot \enc{\dr}(\xx)$, where $\odot$ is the Hadamard (element-wise) product. The prior rates, $\dec{\rr}  \in \RR^K$, are learnable parameters that capture expectations about the statistics of the input. The encoder outputs, $\enc{\dr}(\xx) \in \RR^K$, capture \textit{deviations} from the prior. Thus, \pvae models the interaction between prior expectations, and deviations from them, in a multiplicative and symmetric way. This results in a posterior, $q(\zz \vert \xx) = \pois(\zz; \dec{\rr}\odot\enc{\dr}(\xx))$, and prior, $p(\zz) = \pois(\zz; \dec{\rr})$, where $\zz$ is the spike count variable and $\pois(z; \lambda) = \lambda^z e^{-\lambda}/z!$ is the Poisson distribution. Notably, this multiplicative relationship is maximally general, as any pair of positive variables, $\rr_\text{prior}$, and $\rr_\text{post.}$ can be expressed as a base variable, $\dec{\rr} \coloneqq \rr_\text{prior}$, multiplied by their relative ratio, $\enc{\dr} \coloneqq \rr_\text{post.} / \dec{\rr}$. See \cref{fig:archi+amort}a. 

\paragraph{\pvae loss function.} For a comprehensive derivation of the \pvae objective, see \cref{sec:supp-derivations}. Here, we report the final result:
\begin{empheq}[box=\fbox]{equation}\label{eq:pvae-nelbo}
\begin{aligned}
    \LL_\text{PVAE}
    =
    \expect{
        \zz \sim \pois(\zz; \dec{\rr}\odot\enc{\dr})
        }{
        \norm{\xx - \dec{\mathrm{dec}}(\zz)}_2^2
        }
    +
    \sum_{i=1}^K \dec{r_i} f(\enc{\delta r_i}),
\end{aligned}
\end{empheq}
where $\dec{\mathrm{dec}}(\cdot)$ is the decoder neural network, and $f(y) \coloneqq 1 - y + y\log y$ (see supplementary \cref{fig:poisson-residual}).

\paragraph{\pvae relationship to sparse coding.} The $\mathtt{KL}$ term in \cref{eq:pvae-nelbo} penalizes firing rates. Both $\dec{\rr}$ and $\enc{\dr}$ are positive by definition, and $f(y) \geq 0$, strongly resembling the sparsity penalty in \textcite{olshausen1996emergence}. To make this connection more explicit, we make two additional assumptions (\cref{fig:archi+amort}b):

\begin{enumerate}
    \item The decoder is a linear generative model: $\hat{\xx} = \dec{\bm{\Phi}}\zz$, with $\xx \in \RR^M$ and $\dec{\bm{\Phi}} \in \RR^{M \times K}$.
    \item The latent space is overcomplete: $K > M$.
\end{enumerate}
Because both $\EE_{\zz \sim \pois(\zz; \bm{\lambda})}[z_i]$ and $\EE_{\zz \sim \pois(\zz; \bm{\lambda})}[z_iz_j]$ have closed-form solutions (\cref{eq:supp-expectations}), the reconstruction term in \cref{eq:pvae-nelbo} can be computed analytically for a linear decoder, resulting in:
\begin{empheq}[box=\fbox]{equation}\label{eq:sc-pvae-nelbo}
    \LL_\text{SC-PVAE}\left(
        \xx;
        \enc{\dr},
        \dec{\rr},
        \dec{\bm{\Phi}}
    \right) =
    \norm{\xx - \dec{\bm{\Phi}}\decenc{\bm{\lambda}}}_2^2
    +
    \decenc{\bm{\lambda}}^T\mathrm{diag}(\dec{\bm{\Phi}}^T\dec{\bm{\Phi}})
    +
    \beta \sum_{i=1}^K
        \dec{r_i}
        f(\enc{\delta r_i}).
\end{empheq}
where $\decenc{\bm{\lambda}} = \dec{\rr}\odot\enc{\dr}(\xx)$ are the posterior firing rates, $f(y)$ is defined as above, and $\beta$ is a hyperparameter that scales the contribution of the $\mathtt{KL}$ term \cite{higgins2017beta}, and changes the sparsity penalty for the \pvae.

The relationship between the linear \pvae loss (\cref{eq:sc-pvae-nelbo}) and the sparse coding loss (\cref{eq:sparse-coding}) can now be seen. Both contain a term that minimizes the squared error of the reconstruction and a term (two terms for \pvae) that penalizes non-zero firing rates. Unlike prior work that directly implemented amortized sparse coding \cite{geadah2024sparse, tonolini2020variational}, here the activity penalty naturally emerges from the derivations, and the only additional assumption was an overcomplete linear generative model. The inference is accomplished using a parameterized feed-forward neural network, $\enc{\dr}(\xx)$, thus, it is amortized \cite{ganguly2023amortized}. We call this specific case of \pvae \say{Amortized Sparse Coding} (\cref{fig:archi+amort}b).

Note that a closed-form derivation of the reconstruction term is possible for any VAE with a linear decoder and a generating distribution that has a mean and variance (see \cref{eq:supp-exact-recon-general}).

This closed-form expression of the loss given a linear decoder is useful because we can see how different parameters contribute to the loss. Furthermore, we can compute gradients of the whole loss exactly, and use this to evaluate our Poisson reparameterization.




\begin{table}[t]
    \caption{Models considered in this paper.}
    \label{tab:models}
    \centering
    \begin{tabular}{cc}
        \toprule
        \begin{tabularx}{60mm}{CC}
            \multicolumn{2}{c}{Discrete}\\
            \midrule
            \begin{tabular}{c}
                Poisson VAE \\
                (\pvae)
            \end{tabular}
            &
            \begin{tabular}{c}
                Categorical VAE \\
                (\cvae; \cite{jang2017categorical, maddison2017concrete})
            \end{tabular}
        \end{tabularx}
        &
        \begin{tabularx}{60mm}{CC}
            \multicolumn{2}{c}{Continuous}\\
            \midrule
            \begin{tabular}{c}
                Gaussian VAE \\
                (\gvae; \cite{kingma2014auto, rezende2014stochastic})
            \end{tabular}
            &
            \begin{tabular}{c}
                Laplace VAE \\
                (\lvae; \cite{csikor2023topdown, geadah2024sparse})
            \end{tabular}
        \end{tabularx}
        \\
        \bottomrule
  \end{tabular}
\end{table}

\section{Experiments}\label{sec:experiments}

To evaluate the \pvae, we perform three sets of experiments. First, we utilize the theoretical results for a linear decoder (\cref{eq:sc-pvae-nelbo,eq:supp-exact-recon-general}) to test the effectiveness of our reparameterization algorithm. We compare to alternative VAE models with established reparameterization tricks (e.g., Gaussian).

Second, to confirm \pvae with a linear decoder not only resembles amortized sparse coding but practically performs like sparse coding, we compare to standard and well-established sparse coding algorithms such as the locally competitive algorithm (LCA; \cite{rozell2008sparse}) and the widely-used iterative shrinkage-thresholding algorithm (ISTA; \cite{daubechies2004iterative, fista2009}) to see if \pvae reproduces their results.

Third, we test the \pvae in a generic representation learning context and evaluate the geometry of learned representations for downstream tasks. For these experiments, both the encoder and decoder's architecture is a ResNet (see \cref{sec:supp-architecture-training} for full architecture and training details).

\paragraph{Architecture notation.} We experimented with both convolutional and linear architectures. We highlight the \enc{encoder} and \dec{decoder} networks using \enc{red} and \dec{blue}, respectively. We use the \architecture{enc}{dec} convention to clearly specify which architecture type was used. For example, \convlin represents a model with a convolutional \enc{encoder} and a linear \dec{decoder}. Using this notation, we note that \linlin and \convlin architectures were used for the first and second sets of experiments, while \convconv architectures were employed for the third.

\paragraph{Alternative models.} We compare \pvae to both discrete and continuous VAEs (\Cref{tab:models}). Other than the traditional Gaussian, we compare to Laplace-distributed VAEs because previous work found the Laplace distribution supported robust sparse representations \cite{geadah2024sparse, csikor2023topdown}. Additionally, we compare to Categorical VAEs, trained using the Gumbel-Softmax trick \cite{jang2017categorical, maddison2017concrete}. We use PyTorch's implementation which is based on \textcite{maddison2017concrete}.

Finally, we test models where Gaussian latents are passed through an activation function before passing to the decoder. We call these models \gvae$_\mathrm{\!+act}$, where $\mathrm{act} \in \{\mathrm{relu}, \mathrm{exp}\}$, capturing other families of distributions (truncated Gaussian and log-normal). We include these to test the hypothesis that positive constraints (and not discrete latents) are the key contribution of Poisson \cite{whittington2023disentanglement}.

\paragraph{Datasets.} For sparse coding results, we use $101$ natural images from the van Hateren dataset \cite{van1998independent}. We tile the images to extract $16 \times 16$ patches and apply whitening and contrast normalization, as is typically done in sparse coding literature \cite{olshausen1996emergence, boutin2021sparse}. To test the generalizability of our sparse coding results, we repeat these steps on CIFAR10 \cite{krizhevsky2009learning}, a dataset we call \cifar. For the general representation learning results, we use MNIST. See \cref{sec:supp-architecture-training} for additional details.

\begin{table}[th!]
    \caption{Reparameterized gradient estimators perform comparably to exact ones across datasets and \enc{encoder} architectures (linear vs.~convolutional). Exact gradients are only computable for linear \dec{decoders} (see \cref{eq:supp-exact-recon-general,eq:supp-exact-recon-both,eq:supp-nelbo-poisson-linear}). Values represent percent drop in validation loss (lower is better), shown as \entry{\mathrm{mean}}{99\%} confidence interval calculated from $n=5$ random initializations. The best-performing case was selected as the single best random seed for models of the same architecture and dataset across gradient methods (1 out of: 15 for \pvae, 10 for \gvae). See supplementary \cref{fig:exmcst} for a visualization of the same data presented in this table. For actual loss values, see supplementary \Cref{tab:nelbo-full}. EX: exact; MC: Monte Carlo; ST: straight-through \cite{bengio2013estimating}.}
    \label{tab:exmcst}
    \centering
    \begin{tabular}{l@{\hspace{1mm}}lccc}
        \toprule
        Model
        &
        &
        \begin{tabularx}{34mm}{CC}
            \multicolumn{2}{c}{van Hateren}\\
            \midrule
            {\small\linlin} & {\small\convlin}
        \end{tabularx}
        &
        \begin{tabularx}{34mm}{CC}
            \multicolumn{2}{c}{\cifar}\\
            \midrule
            {\small\linlin} & {\small\convlin}
        \end{tabularx}
        &
        \begin{tabularx}{34mm}{CC}
            \multicolumn{2}{c}{MNIST}\\
            \midrule
            {\small\linlin} & {\small\convlin}
        \end{tabularx}
        \\
        \midrule
        \pvae
        &
        \begin{tabularx}{5mm}{C}
            EX\\MC\\ST
        \end{tabularx}
        &
        \begin{tabularx}{34mm}{CC}
            \entry{0.6}{.5} & \entry{0.1}{.1} \\ 
            \entry{0.0}{.1} & \entry{0.7}{.1} \\ 
            \entry{7.3}{.1} & \entry{10.5}{.1}
        \end{tabularx}
        &
        \begin{tabularx}{34mm}{CC}
            \entry{0.0}{.1} & \entry{0.0}{.0} \\ 
            \entry{0.2}{.0} & \entry{0.5}{.1} \\ 
            \entry{9.1}{.1} & \entry{12.5}{.1}
        \end{tabularx}
        &
        \begin{tabularx}{34mm}{CC}
            \entry{0.1}{.1} & \entry{0.5}{.6} \\ 
            \entry{0.7}{.4} & \entry{0.9}{.5} \\ 
            \entry{8.1}{.3} & \entry{11.8}{.2}
        \end{tabularx}
        \\
        \midrule
        \gvae
        &
        \begin{tabularx}{5mm}{C}
            EX\\MC
        \end{tabularx}
        &
        \begin{tabularx}{34mm}{CC}
            \entry{0.1}{.1} & \entry{0.0}{.0} \\ 
            \entry{0.1}{.1} & \entry{0.0}{.0}
        \end{tabularx}
        &
        \begin{tabularx}{34mm}{CC}
            \entry{0.0}{.1} & \entry{0.0}{.0} \\ 
            \entry{0.1}{.1} & \entry{0.0}{.0}
        \end{tabularx}
        &
        \begin{tabularx}{34mm}{CC}
            \entry{0.1}{.2} & \entry{0.1}{.2} \\ 
            \entry{0.4}{.1} & \entry{0.3}{.1}
        \end{tabularx}
        \\
        \bottomrule
  \end{tabular}
\end{table}

\paragraph{Statistical tests.} In the VAE literature, it is known that random seeds can have a large effect compared to architecture or regularization \cite{locatello2019challenging}. Therefore, we train each configuration using 5 different random initializations. We report $99\%$ confidence intervals throughout, and perform paired $t$-tests, reporting significance for $p < 0.01$ (FDR corrected using the Benjamini-Hochberg method).



\paragraph{Evaluating the Poisson reparameterization algorithm.} \pvae with a linear decoder has a closed form solution (\cref{eq:sc-pvae-nelbo}), which lets us evaluate how well our reparameterized gradients perform compared to the exact ones. We compare our results to the gold-standard Gaussian (\Cref{tab:exmcst}), as well as Categorical and Laplace VAEs (supplementary \Cref{tab:nelbo-full}). In \Cref{tab:exmcst}, we report the percent performance drop relative to the best fit, enabling meaningful comparisons across architectures and datasets. Monte Carlo sampling with Poisson reparameterization closely matches exact inference just like established methods for Gaussian and Laplace. In contrast, the straight-through (ST; \cite{bengio2013estimating}) estimator performs poorly (\Cref{tab:exmcst}; see also supplementary \cref{fig:exmcst}).

\paragraph{Annealing the temperature.} The temperature parameter ($T$) is a crucial hyperparameter in our Poisson reparameterization trick (\Cref{algo:rsample}). To assess its impact, we followed standard practice \cite{jang2017categorical} and annealed $T$ during the first half of training, starting from a large value ($T_\text{start} = 1$) and gradually decreasing it to a small value ($T_\text{final} = 0.05$ in the main paper). \Cref{fig:hard-fwd} shows the performance on the van Hateren dataset as a function of various $T_\text{final}$, two architectures (\linlin and \convlin), as well as two annealing schedules (linear vs.~exponential; see inset). We find that final temperatures $T_\text{final} \leq 0.1$ and either annealing strategy work well.

During training, we maintain $T > 0$, which results in continuous (floating) latent variables, $\zz$. At test time, we set $T = 0$ to produce genuine integer Poisson samples. Crucially, all reported results use $T = 0$ at test time. We also explored a \say{hard-forward} scheme during the latter half of training, where $T$ remains nonzero only in the backward pass. This \textit{surrogate gradients} approach provides integer latents in the forward pass but, somewhat unexpectedly, underperformed our \say{relaxed Poisson} method (\cref{fig:hard-fwd}). These findings suggest that surrogate gradient methods might benefit from relaxing the hard-forward strategy during training. We believe this observation will be of particular interest to the spiking neural network community, which often relies on surrogate gradients for training.

\begin{figure}[t!]
    \centering
    \includegraphics[width=1.0\linewidth]{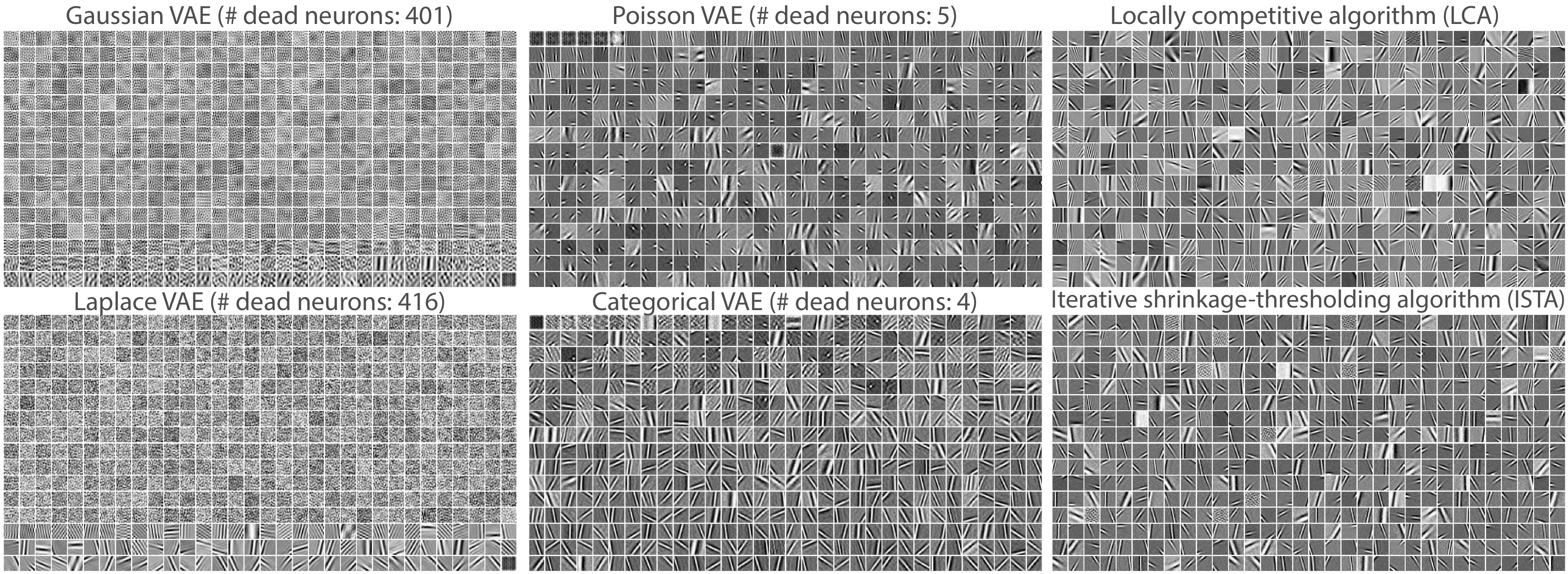}
    \caption{Learned basis elements for various \linlin VAEs (first two columns) and standard sparse coding models (last column). There are a total of $K=512$ elements, each made of $16 \times\! 16 = 256$ pixels (i.e., $\dec{\bm{\Phi}} \in \RR^{256 \times 512}$). Features are ordered from top-left to bottom-right, in ascending order of their associated $\mathtt{KL}$ divergence (\pvae, \gvae, \lvae), or the magnitude of posterior $\mathrm{logits}$ (\cvae). The sparse coding results (LCA and ISTA) are ordered randomly.}
    \label{fig:phi}
\end{figure}

\paragraph{The \pvae learns basis vectors similar to those from sparse coding.} A major result from sparse coding is that it learns basis vectors (dictionaries) that resemble the \say{Gabor-like} receptive fields of cortical neurons \cite{olshausen1996emergence, Hubel1959ReceptiveFO, Hubel1968ReceptiveFA}. Inspecting the dictionaries learned by different models demonstrates this is not trivial (\cref{fig:phi}). As expected from theoretical results \cite{tipping1999prob}, \gvae (top left) learn probabilistic PCA, but with many noisy elements. As demonstrated previously \cite{geadah2024sparse, csikor2023topdown}, \lvae (lower left) learn Gabor-like elements. However, there are a large number of noisy basis vectors. It is of note that previous work did not show complete dictionaries for their results with Laplace latents \cite{geadah2024sparse, csikor2023topdown}. In contrast, \pvae (top middle) learns Gabor-like filters that cover space, orientation, and spatial frequency. The quality is comparable to sparse coding dictionaries learned with LCA/ISTA (top/lower right panels). \cvae also learns Gabors, although there are significantly more noisy basis elements.

\paragraph{The \pvae avoids posterior collapse.} A striking feature of \cref{fig:phi} is the sheer number of noisy basis vectors for both continuous VAEs (\gvae, \lvae). We suspected this reflected dead neurons with vanishing $\mathtt{KL}$, which is indicative of a collapsed latent dimension that's no longer encoding information. To quantify this, we binned the distribution of $\mathtt{KL}$ values and thresholded the resulting distribution at discontinuous points (see supplemental \cref{fig:dead-kl}). \Cref{tab:active} shows the results of this analysis for all VAEs with valid $\mathtt{KL}$ terms. Across all datasets, both continuous VAEs suffered from large numbers of dead neurons, whereas \pvae largely avoided this problem. On both natural image datasets, \pvae had $\sim\! 2\%$ dead neurons compared to $\sim\! 80\%$  for \gvae and \lvae. Having a more expressive encoder slightly increases this percentage, but a dramatic difference between \pvae and continuous VAEs (\gvae, \lvae) persists.

\paragraph{The \pvae learns sparse representations.} To quantify whether \pvae learns sparse representations, we compared our VAE models to sparse coding trained with LCA and ISTA and quantified the lifetime sparsity \cite{vinje2000sparse}. The lifetime sparsity of the $j$-th latent is:
\begin{equation}\label{eq:lifetime}
    s_j = \left(1 - \frac{1}{N}\right)^{-1} \left(1 - \frac{1}{N} \frac{\left(\sum_i z_{ij}\right)^2}{\sum_i z_{ij}^2}\right),
\end{equation}
where $N$ is the number of images, and $z_{ij}$ is sampled from the posterior for the $i$-th image. Intuitively, $s_j = 1$ whenever neuron $j$ responds to a single stimulus out of the entire set (highly selective). In contrast, $s_j = 0$ whenever the neuron responds equally well to all stimuli indiscriminately.

\begin{figure}[ht!]
    \centering
    \includegraphics[width=\linewidth]{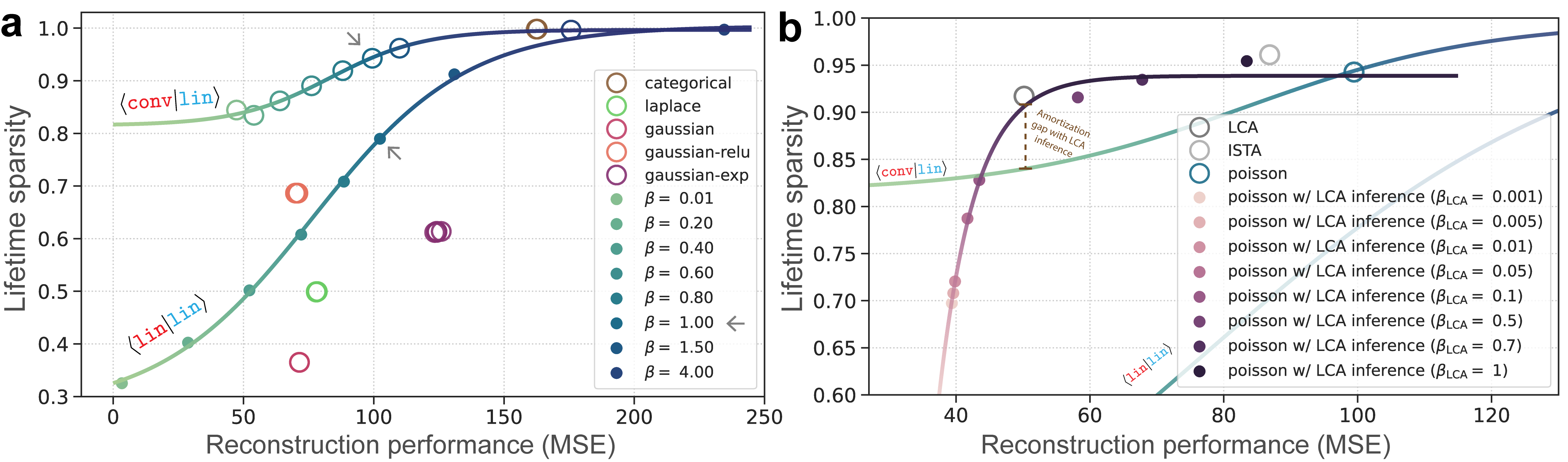}
    \caption{
        Reconstruction performance vs. sparsity of representations. \textbf{(a)} Results for the VAE model family. The curves are sigmoid fit to \linlin and \convlin \pvae results across varying $\beta$ values ($\beta$ from \cref{eq:sc-pvae-nelbo}). Empty circles correspond to \convlin architectures. \textbf{(b)} Amortization gap for \pvae (blue open circle) compared to sparse coding (LCA/ISTA). Solid points show results from applying the LCA inference algorithm to \pvae basis vectors at different sparsity levels ($\beta_\text{LCA}$ from \cref{eq:sparse-coding}). The purple curve is a sigmoid fit, and curves from part (a) are also included for comparison.
    }
    \label{fig:rate-dist}
\end{figure}

\cref{fig:rate-dist}a shows the reconstruction performance ($\mathrm{MSE}$) compared to lifetime sparsity ($s$, \cref{eq:lifetime}) for all VAEs. Empty and solid circles represent \convlin and \linlin architectures, respectively. The \gvae finds good reconstructions ($\mathrm{MSE} = 71.49$) but with low sparsity ($s = 0.37$).
Because the \pvae $\mathtt{KL}$ term explicitly penalizes rate (\cref{eq:pvae-nelbo}), we explored different $\beta$ values for \pvae with both \linlin and \convlin architectures (\cref{fig:rate-dist}a, blue curves). This maps out rate-distortion curves, enabling us to compare the sparsity levels at which \pvae matches \gvae performance.

With a simpler (linear) encoder, \linlin \pvae matches \convlin \gvae performance while achieving $1.7\times$ greater sparsity at $\beta = 0.6$. A \convlin \pvae further increases this gap to $2.4\times$ greater sparsity. Adding a $\mathrm{relu}$ activation to \gvae also increases sparsity ($s = 0.69$). By comparing \linlin and \convlin \pvae models, we observe that enhancing encoder complexity for the same $\beta = 1$ (gray arrows) preserves $\mathrm{MSE}$ performance while achieving greater sparsity. This highlights how amortization quality can significantly influence rate-distortion curves \cite{alemi18fixing, cremer2018inference, marino18iterative, vahdat2020nvae}.

\begin{table}[t!]
    \caption{Proportion of active neurons. All models considered in this table had a latent dimensionality of $K=512$, with either \linlin or \convlin architectures. See also supplementary \cref{fig:dead-kl}.}
    \label{tab:active}
    \centering
    \begin{tabular}{l@{\hspace{1mm}}ccc}
        \toprule
        \hspace{2mm}Model
        &
        \begin{tabularx}{37mm}{CC}
            \multicolumn{2}{c}{van Hateren}\\
            \midrule
            linear & conv
        \end{tabularx}
        &
        \begin{tabularx}{37mm}{CC}
            \multicolumn{2}{c}{\cifar}\\
            \midrule
            linear & conv
        \end{tabularx}
        &
        \begin{tabularx}{37mm}{CC}
            \multicolumn{2}{c}{MNIST}\\
            \midrule
            linear & conv
        \end{tabularx}
        \\
        \midrule
        \begin{tabular}{l}
            \pvae \\
            \lvae \\
            \gvae
        \end{tabular}
        &
        \begin{tabularx}{37mm}{CC}
            \entry{\bf0.984}{.011} & \entry{\bf0.819}{.041} \\ 
            \entry{0.188}{.000} & \entry{0.222}{.003} \\ 
            \entry{0.218}{.003} & \entry{0.246}{.000}
        \end{tabularx}
        &
        \begin{tabularx}{37mm}{CC}
            \entry{\bf0.999}{.002} & \entry{\bf0.928}{.045} \\ 
            \entry{0.193}{.003} & \entry{0.230}{.000} \\ 
            \entry{0.105}{.008} & \entry{0.246}{.000}
        \end{tabularx}
        &
        \begin{tabularx}{37mm}{CC}
            \entry{\bf0.537}{.008} & \entry{\bf0.426}{.011} \\ 
            \entry{0.027}{.000} & \entry{0.034}{.002} \\ 
            \entry{0.027}{.000} & \entry{0.031}{.000}
        \end{tabularx}
        \\
        \bottomrule
  \end{tabular}
\end{table}

Does \pvae match the performance of traditional sparse coding trained with LCA or ISTA? \Cref{fig:rate-dist}b compares \pvae to sparse coding models that were trained using a wide range of hyperparameters, and the best models were selected for each class (\cref{sec:supp-architecture-training}). \pvae achieves a similar sparsity to LCA and ISTA ($s = 0.94, 0.91$, and $0.96$, respectively), but the best LCA model drastically outperforms \pvae on $\mathrm{MSE}$ for similar levels of sparsity. This suggests our convolutional encoder is struggling to close the amortization gap. To test this hypothesis, we performed LCA inference on basis elements learned by \pvae (\cref{fig:rate-dist}b curve/solid points). We explored a range of hyperparameters to determine whether the $\mathrm{MSE}$ improved for similar sparsity levels. Indeed, LCA inference using \pvae dictionary was able to nearly match the performance of sparse coding LCA for similar levels of sparsity. This confirms our hypothesis that a large amortization gap remains for the specific encoder architectures we tested, highlighting the need for improved inference algorithms/architectures \cite{cremer2018inference}.

\paragraph{The \pvae is more sample efficient in downstream tasks.} To assess downstream performance, we trained \convconv VAE models with a $K=10$ latent dimension on MNIST (see supplementary \cref{fig:recon+sample} for generated samples and reconstructions from these models). We then extracted representations from the trained encoders and evaluated their ability to classify MNIST digits. We define representations as mean vectors $\bmu$ for continuous VAEs (\gvae, \lvae) following conventions in the VAE literature \cite{locatello2019challenging}, and use $\log \enc{\dr}$ for \pvae, and logits for \cvae.

We split the MNIST validation set into two $5{,}000$ sample sets, used as train/test sets for this task. We train K-nearest neighbors (KNN) classifiers with a varying number of limited supervised samples ($N = 200, 1000, 5000$) drawn without replacement from the first set (train), to measure classification accuracy on the withheld set (test). KNN is nonparametric, and its performance is directly influenced by the geometry of representations by explicitly capturing the distance between encoded samples \cite{weinberger2009distance}. We find that using only $N=200$ samples, \pvae achieves $\sim\! 82\%$ accuracy in held out data; whereas, \gvae achieves the same level of accuracy at $N=1000$ samples (\Cref{tab:geom}). By this measure, \pvae is $5\times$ more sample efficient. But from \textcite{alleman2024task}, we know that the choice of activation function changes the geometry of learned representations. Therefore, we also tested \gvae models with an activation function ($\mathrm{relu}$ and $\mathrm{exp}$) applied to latents after sampling from the posterior. This biological constraint improved \gvae, but it still underperformed \pvae (\Cref{tab:geom}). We also found this result held for higher dimensional latent spaces (supplementary \Cref{tab:geom-full}).

In supplementary analyses (\cref{fig:logreg}), we evaluated the representations using logistic regression trained on the full dataset. For larger latent dimensionalities ($K=50, 100$), \pvae outperformed all other VAEs, but at lower dimensionalities ($K=10$), it underperforms both \gvae and \lvae.

\paragraph{The \pvae learns representations with higher dimensional geometry.} The preceding results are indicative of substantial differences in the geometry of the representations learned by \pvae compared to other VAE families (\Cref{tab:geom}). To test this more explicitly, we calculated the \say{shattering dimensionality} of the latent space \cite{rigotti2013importance, bernardi2020geometry, kaufman2022implications}. Shattering dim measures the average accuracy over all possible pairwise classification tasks. This is called \say{shattering} because if the model shatters data points around into a high dimensional space, they will become more linearly separable. For MNIST with 10 classes, there are $\binom{10}{5} = 252$ possible classifications. We trained logistic regression on the entire training set to classify each of the 252 arbitrary splits and measured the average performance on the entire validation set. The far right column of \Cref{tab:geom} shows the measured shattering dims. For $K = 10$, the shattering dim was significantly higher for discrete VAEs (\pvae, \cvae). For higher dimensional latent spaces \pvae strongly outperformed alternative models (\Cref{tab:geom-full}).

\begin{table}[t]
    \caption{Geometry of representations ($K = 10$ only; see \Cref{tab:geom-full} for the full set of results).}
    \label{tab:geom}
    \centering
    \begin{tabular}{clcc}
        \toprule
        \begin{tabular}{c}
            Latent\\dim.
        \end{tabular}
        &
        \hspace{2.5mm}Model
        &
        \begin{tabularx}{65mm}{CCC}
            \multicolumn{3}{c}{KNN classification ($N,$ \# labeled samples)}\\
            \midrule
            $N=200$ & $N=1{,}000$ & $N=5{,}000$
        \end{tabularx}
        &
        \begin{tabular}{c}
            Shattering\\dim.
        \end{tabular}
        \\
        \midrule
        $K = 10$
        &
        \begin{tabular}{l}
            \pvae \\
            \cvae \\
            \lvae \\
            \gvae \\
            \grelu \\
            \gexp
        \end{tabular}
        &
        \begin{tabularx}{65mm}{CCC}
            \entry{\bf0.815}{.002} & \entry{\bf0.919}{.001} & \entry{\bf0.946}{.017} \\ 
            \entry{0.705}{.002} & \entry{0.800}{.002} & \entry{0.853}{.040} \\ 
            \entry{0.757}{.003} & \entry{0.869}{.002} & \entry{\bf0.924}{.028} \\ 
            \entry{0.673}{.003} & \entry{0.813}{.002} & \entry{0.891}{.033} \\ 
            \entry{0.694}{.003} & \entry{0.817}{.003} & \entry{0.877}{.045} \\ 
            \entry{0.642}{.003} & \entry{0.784}{.002} & \entry{0.863}{.032}
        \end{tabularx}
        &
        \begin{tabular}{c}
            \entry{\bf0.797}{.009} \\ 
            \entry{\bf0.795}{.006} \\ 
            \entry{0.751}{.008} \\ 
            \entry{0.758}{.007} \\ 
            \entry{0.762}{.007} \\ 
            \entry{0.737}{.008}
        \end{tabular}
        \\
        \bottomrule
  \end{tabular}
\end{table}

\section{Conclusions}\label{sec:conclusion}

In this paper, we introduced the \pvae, a generative model that encodes inputs into discrete spike counts and unifies established theoretical concepts in neuroscience with modern machine learning. We introduced a Poisson reparameterization algorithm and derived the ELBO for Poisson-distributed latent variables. The \pvae objective results in a $\mathtt{KL}$ term that penalizes firing rates, like sparse coding. We showed that \pvae with a linear decoder reduces to amortized sparse coding. We evaluated the representations on downstream classification tasks and found that \pvae encodes its inputs in a higher dimensional space, enabling good linear separability between classes.

\paragraph{Limitations.}\label{sec:limitations} \pvae samples Poisson latents. Although this is inspired by the statistics of spike counts in the brain over short time intervals \cite{teich1989fractal}, there are deviations from Poisson throughout the cortex over longer time windows \cite{goris2014partitioning}. We discuss this point in \cref{sec:poissonassumption}. A second limitation is the amortization gap between our current implementation of \pvae and traditional sparse coding. This could likely be closed with more expressive encoders \cite{gregor2010learning} or through iterative inference \cite{marino18iterative, kim18savae}, but it is an open area of research \cite{cremer2018inference}.

\paragraph{Neuroscience implications and future directions.} Like biological neurons, the P-VAE generates spikes. This non-negative, discrete representational form closely parallels neuronal spiking activity. Therefore, the \pvae can be more directly compared to neuronal circuits than unconstrained, continuous VAEs. This analogy facilitates in silico perturbation experiments (e.g., \say{stimulating} or \say{silencing} P-VAE neurons) to mirror in vivo causal manipulations. It also allows applying methods like \textit{Most Exciting Inputs} (MEI; \cite{walker2019inception}), which assume non-negative activations. Future work could explore hierarchical P-VAEs, finding a sweet spot between interpretability and performance. Overall, the biologically inspired representational form of P-VAE brings computational modeling closer to experimental neuroscience and opens new avenues for advancing NeuroAI research \cite{zador2023catalyzing, doerig2023neuroconnectionist}.

\section{Code and data}
Our code, data, and model checkpoints are available here: \href{https://github.com/hadivafaii/PoissonVAE}{\color{purple}https://github.com/hadivafaii/PoissonVAE}.

\section{Acknowledgments}
This work was supported by the National Institute of Health under award number NEI EY032179. Additionally, this material is based upon work supported by the National Science Foundation Graduate Research Fellowship Program under Grant No.~DGE-1752814 (DG). Any opinions, findings, conclusions, or recommendations expressed in this material are those of the author(s) and do not necessarily reflect the views of the National Science Foundation. We thank our anonymous reviewers for their helpful comments, and the developers of the software packages used in this project, including PyTorch \cite{paszke2019pytorch}, NumPy \cite{harris2020array}, SciPy \cite{SciPy2020}, scikit-learn \cite{pedregosa2011scikit}, pandas \cite{reback2020pandas}, matplotlib \cite{hunter2007matplotlib}, and seaborn \cite{waskom2021seaborn}.

\printbibliography
\clearpage


\appendix

\section{Are real neurons truly Poisson?}
\label{sec:poissonassumption}
In this section, we discuss empirical and theoretical observations from neuroscience that motivated our Poisson assumption.

\say{Poisson-like} noise in neuroscience has a long history. It begins with observations that neurons do not fire the same sequence of spikes to repeated presentations of the same input, and that the variance is proportional to the mean \cite{tolhurst1983statistical, dean1981variability}, and was followed by the observation that for short counting windows, that proportionality is one \cite{teich1989fractal, shadlen1998variable, averbeck2006neural, rieke1999spikes, abbott2001theoretical}. Larger windows and higher visual areas are notably super-Poisson, but that can be attributed to a modulation of the rate of an inhomogeneous Poisson process \cite{goris2014partitioning}.

In other words, neurons are conditionally Poisson, not marginally Poisson \cite{truccolo2005point}.

Spike-generation, it is argued, is not noisy \cite{mainen1995reliability, deweese2003binary, calvin1968synaptic}, but synaptic noise \cite{allen1994evaluation}, or noise on the membrane potential, can create a Poisson-like distribution of spikes \cite{carandini2004amplification}. An important caveat is that the well-known example of precision in spike generation by \textcite{mainen1995reliability} is effectively captured by a Poisson-process Generalized Linear Model (GLM; \textcite{weber2017capturing}). However, this precision relies on a Bernoulli approximation to a Poisson process, allowing only 0 or 1 spikes. There is a widely-held misconception that precise timing cannot be produced by spike-rate models, but inhomogeneous rate models can produce precise spiking patterns at high time resolution \cite{butts2016nonlinear}. In contrast, recent work has shown that correlations in excitatory inputs drive Poisson-like variability, explaining the widespread observation of Poisson-like noise in real neurons \cite{pattadkal2024synchrony}.

In summary, neurons are not literally Poisson, but it is a good choice. To set up the ELBO, one has to choose an approximate posterior and prior. Because spike counts are integer and cannot be negative, Poisson is a more natural choice than Gaussian without knowing anything about neural firing statistics. Here, we found that the Poisson assumption led to a model with interesting theoretical and empirical properties, where sparse coding emerged from the ELBO with Poisson.

Extending the \pvae to hierarchical architectures \cite{sonderby2016ladder, vahdat2020nvae, child2021vdvae, vafaii2023hierarchical} will make the latents conditionally Poisson, but not marginally Poisson (as they are modulated by top-down rates). Further extensions could implement doubly-stochastic spike generation \cite{goris2014partitioning, aghamohammadi2024doubly}.

\section{Full derivations}\label{sec:supp-derivations}
In this section, we provide a self-contained and pedagogical introduction to VAEs, derive the \pvae loss function, and highlight how combining Poisson-distributed latents with predictive coding leads to the emergence of a metabolic cost term in the \pvae loss. For the case of a linear decoder, the reconstruction loss assumes a closed-form solution. This means we can compute the gradients analytically, which we can then use to evaluate the Poisson reparameterization trick.

\subsection{Deriving the evidence lower bound (ELBO) loss}
For completeness, let's first go over the basics. This section will provide a quick refresher on variational inference and how to derive the VAE loss from scratch. Assume the data $\xx \in \RR^M$ and $K$-dimensional latent variables $\zz$ are jointly distributed as $p(\xx, \zz)$, with the data generated through the following process:
\begin{equation}
    p(\xx) = \int p(\xx, \zz) \, d\zz = \int p(\xx \vert \zz) p(\zz) \, d\zz,
\end{equation}

In Bayesian posterior inference, the goal is to identify which latents $\zz$ are likely given data $\xx$. In other words, we want to approximate $P(\zz \vert \xx)$, the optimal but (typically) intractable posterior distribution.

\subsubsection{Variational inference and VAE loss function} To achieve approximate Bayesian inference, a common approach is to define a family of variational densities $\mathcal{Q}$ and find a member $q(\zz \vert \xx) \in \mathcal{Q}$ such that it sufficiently approximates the optimal posterior \cite{blei2017variational}. We call $q(\zz \vert \xx)$ the \textit{approximate posterior}. The general aim of variational inference ($\mathtt{VI}$) can be summarized as follows:
\begin{equation}\label{eq:supp-vi}
    \mathtt{VI}: \quad \text{find a } q(\zz \vert \xx) \in \mathcal{Q} \text{ such that } q(\zz \vert \xx) \text{ is a good approximation of } p(\zz \vert \xx).
\end{equation}

The goodness of our approximate posterior, or its closeness to the true posterior, is measured using the Kullback-Leibler ($\mathtt{KL}$) divergence:

\begin{equation}\label{eq:supp-kl-def}
q^* = \argmin_{q \, \in \mathcal{Q}} \KL{q(\zz \vert \xx)}{p(\zz \vert \xx)}.
\end{equation}

We cannot directly optimize \cref{eq:supp-kl-def}, because $p(\zz \vert \xx)$ is often intractable. Instead, we rearrange some terms and arrive at the following loss function:

\begin{equation}\label{eq:supp-nelbo}
\LL_\text{NELBO}(q) = - \expect{\zz \sim q(\zz \vert \xx)}{\log p(\xx \vert \zz)} + \KL{q(\zz \vert \xx)}{p(\zz)}.
\end{equation}

NELBO stands for negative ELBO, also known as \say{variational free energy.} Notably, finding a $q \in \mathcal{Q}$ that minimizes $\LL_\text{NELBO}(q)$ in \cref{eq:supp-nelbo} is equivalent to finding the optimal $q^*$ in \cref{eq:supp-kl-def}.

The first term in \cref{eq:supp-nelbo}, often called the reconstruction term, captures the likelihood of the observed data $\xx$, given latents $\zz$, under the approximate posterior. For all our VAE models, we assume a Gaussian conditional likelihood with a fixed variance, as is typically done in the literature. This approximates the reconstruction term as the mean squared error between input data and their reconstructed version. The second term, known as the $\mathtt{KL}$ term, is more interesting. This term can assume very different forms depending on the distribution used.

\subsection{The KL term}

In this section, we will derive closed-form expressions for the $\mathtt{KL}$ term for different choices of the distributions $q(\zz \vert \xx)$ and $p(\zz)$. Specifically, we will focus on Gaussian and Poisson parameterizations.

\paragraph{Predictive coding assumption.} We will draw inspiration from predictive coding and assume that the bottom-up inference pathway only encodes the residual information relative to the top-down, or predicted information. We will apply this idea to both Gaussian and Poisson cases, and find that only in the Poisson case, the outcome becomes interpretable and resembles sparse coding objective.

\subsubsection{KL term derivation: Gaussian}

Let $q(\zz \vert \xx) = \NN(\zz; \enc{\bmu_q}(\xx), \enc{\bsig_q}(\xx))$ and $p(\zz) = \NN(\zz; \dec{\bmu_p}, \dec{\bsig_p})$, where the mean and variance are either outputs of the \enc{encoder} network or parameters of the \dec{decoder} network.

Now, let us implement the predictive coding assumption, where the encoder only keeps track of residual information that is not already contained in the prior information. Mathematically, this idea can be formalized as follows:

\begin{equation}
\begin{aligned}
    \dec{\bmu_p}
    &\rightarrow
    \dec{\bmu},
    \quad\quad
    \enc{\bmu_q}
    \rightarrow
    \dec{\bmu} +
    \enc{\bm{\delta\mu}}
    \\
    \dec{\bsig_p}
    &\rightarrow
    \dec{\bsig},
    \quad\quad
    \enc{\bsig_q}
    \rightarrow
    \dec{\bsig} \cdot 
    \enc{\bm{\delta\sigma}}
\end{aligned}
\end{equation}

With these modifications, the Gaussians $\mathtt{KL}$ term becomes:

\begin{equation}\label{eq:supp-kl-gaussian}
    \kl{\enc{q}}{\dec{p}} =
    \frac{1}{2} \Big(
        \frac{
            \enc{\bm{\delta\mu}}^2}{
            \dec{\bsig}^2
        }
        + \enc{\bm{\delta\sigma}}^2
        - \log\enc{\bm{\delta\sigma}}^2
        - \bm{1}
    \Big).
\end{equation}

In standard Gaussian VAEs, the prior has no learnable parameter. Instead, we have $\dec{\bmu} \rightarrow \bm{0}$ and $\dec{\bsig} \rightarrow \bm{1}$. Therefore, the final form of the $\mathtt{KL}$ term for a standard Gaussian VAE is:

\begin{equation}\label{eq:supp-kl-gaussian-standard}
    \kl{\enc{q}}{\NN(\bm{0}, \bm{1})} =
    \frac{1}{2} \Big(
        \enc{\bm{\delta\mu}}^2
        + \enc{\bm{\delta\sigma}}^2
        - \log\enc{\bm{\delta\sigma}}^2
        - \bm{1}
    \Big).
\end{equation}

We observe that the $\mathtt{KL}$ term vanishes when $\enc{\bm{\delta\mu}} \rightarrow \bm{0}$ and $\enc{\bm{\delta\sigma}} \rightarrow \bm{1}$. This happens whenever no new information is propagated through the encoder, a phenomenon known as posterior collapse.

Other than this trivial observation, \cref{eq:supp-kl-gaussian-standard} does not really lend itself to interpretation. In contrast, will show below that a Poisson parameterization of VAEs leads to a much more interpretable outcome for the $\mathtt{KL}$ term.

\subsubsection{KL term derivation: Poisson}

Now suppose $q(z \vert \xx) = \pois(z; \dec{r}\enc{\delta r}(\xx))$, and $p(z) = \pois(z; \dec{r})$, where $z$ is literally the spike count of a single latent dimension---or shall we say, neuron?

In the Poisson case, the $\mathtt{KL}$ term becomes more interpretable, as we will show below. Recall that the Poisson distribution for a single variable $z$, given rate $\lambda \in \RR_{>0}$, is given by:

\begin{equation}\label{eq:pois}
    \pois(z; \lambda) = \frac{\lambda^{z}e^{-\lambda}}{z!}.
\end{equation}

Plug this expressions into the $\mathtt{KL}$ divergence definition to get:

\begin{equation}\label{eq:supp-kl-poisson}
\begin{aligned}
    \kl{\enc{q}}{\dec{p}}
    &=
    \expect{z \sim \enc{q}}{\log \frac{\enc{q}}{\dec{p}}}
    \\
    &=
    \EXPECT{z \sim \enc{q}}{
        \log
        \frac{(\dec{r}\enc{\delta r})^z
        e^{-\dec{r}\enc{\delta r}}
        /z!
        }{
        \dec{r}^z
        e^{-\dec{r}}
        /z!
        }
    }
    \\
    &=
    \EXPECT{z \sim \enc{q}}{
        \log
        \left(
            \left(
            \frac{\dec{r}\enc{\delta r}}{\dec{r}}
            \right)^z
            e^{-\dec{r}\enc{\delta r} + \dec{r}}
        \right)
    }
    \\
    &=
    \EXPECT{z \sim \enc{q}}{
        \log
        \enc{\delta r}^z
        +
        \log
        e^{-\dec{r}\enc{\delta r} + \dec{r}}
    }
    \\
    &=
    \EXPECT{z \sim \enc{q}}{
        z\log\enc{\delta r}
        -
        \dec{r}\enc{\delta r}
        +
        \dec{r}
    }
    \\
    &=
    \expect{z \sim \enc{q}}{z}
    \log\enc{\delta r}
    -
    \dec{r}\enc{\delta r}
    +
    \dec{r}
    \\
    &=
    \dec{r}\enc{\delta r}
    \log\enc{\delta r}
    -
    \dec{r}\enc{\delta r}
    +
    \dec{r}
    \\
    &=
    \dec{r} \left(
        1 - \enc{\delta r} +
        \enc{\delta r} \log \enc{\delta r}
    \right)
    \\
    &=
    \dec{r} f(\enc{\delta r}),
\end{aligned}
\end{equation}
where we have define $f(y) \coloneqq 1 - y + y\log y$. 

To examine the behavior of the Poisson $\mathtt{KL}$ term, we assume $\enc{\delta r} = 1 + \enc{\eps}$, where $\enc{\eps} \ll 1$, then Taylor expand $f$. Calculating the first and second derivatives of $f(y) = 1 - y + y\log y$ gives $f'(y) = \log y$ and $f''(y) = 1/y$. Thus:
\begin{equation}\label{eq:supp-f-quadratic}
\begin{aligned}
    f(1 + \enc{\eps})
    &=
    f(1)
    +
    \enc{\eps} f'(1)
    +
    \frac{\enc{\eps}^2}{2!} f''(1)
    +
    \mathcal{O}(\enc{\eps}^3)
    \\
    &=
    0 + 0
    + \frac{\enc{\eps}^2}{2!}
    + \mathcal{O}(\enc{\eps}^3)
    \\
    &\approx
    \frac{1}{2}\enc{\eps}^2
\end{aligned}
\end{equation}

Plug this back into \cref{eq:supp-kl-poisson} to get:
\begin{equation}
\begin{aligned}
    \kl{\enc{q}}{\dec{p}}
    &=
    \dec{r} f(\enc{\delta r})
    \\
    &=
    \dec{r} f(1 + \enc{\eps})
    \\
    &\approx
    \frac{1}{2}\dec{r}\enc{\eps}^2.
\end{aligned}
\end{equation}

\begin{figure}[ht!]
    \centering
    \includegraphics[width=\linewidth]{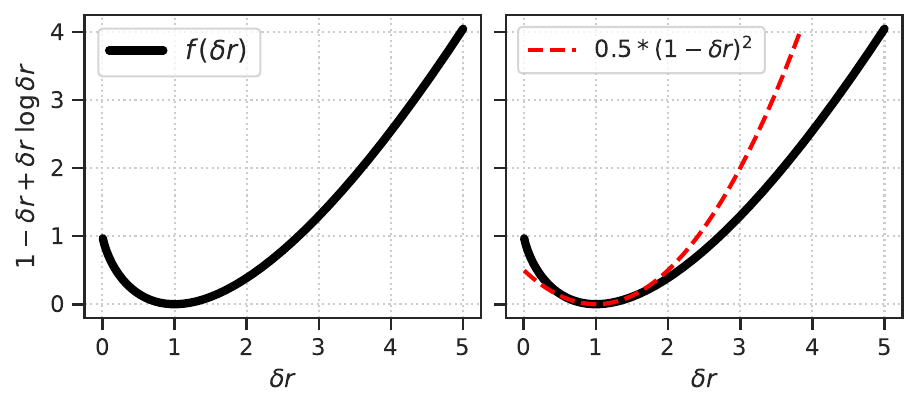}
    \caption{Left, residual term $f(\delta r)$ from \cref{eq:supp-kl-poisson}. Right, quadratic approximation of $f$ from \cref{eq:supp-f-quadratic}.}
    \label{fig:poisson-residual}
\end{figure}

For small deviations $\enc{\eps}$, the $\mathtt{KL}$ term simplifies to the product of the prior firing rate, $\dec{r}$, and $\enc{\eps}^2$. See \cref{fig:poisson-residual} for a visualization of the full function, $f(\enc{\delta r}) = 1 - \enc{\delta r} + \enc{\delta r} \log \enc{\delta r}$, along with its quadratic approximation near $\enc{\delta r} = 1$.

In general, there are two ways to minimize the $\mathtt{KL}$ term: dead prior neurons ($\dec{r} \rightarrow 0$), or posterior collapse ($\enc{\delta r} \rightarrow 1$).

Together with the reconstruction loss, the NELBO for a 1-dimensional \pvae reads:

\begin{equation}\label{eq:supp-nelbo-poisson}
    \LL_\text{PVAE}
    \left(
        \dec{r},
        \enc{\delta r}
    \right)
    =
    \LL_\mathrm{recon.}\left(
        \dec{r},
        \enc{\delta r}
    \right)
    +
    \dec{r} \left(
        1 - \enc{\delta r} +
        \enc{\delta r} \log \enc{\delta r}
    \right).
\end{equation}

Finally, it is easy to show that for $K$-dimensional latent space, \cref{eq:supp-kl-poisson} generalizes to:

\begin{equation}
    \KL{
        \pois(\zz; \dec{\rr} \odot \enc{\dr}(\xx))
        }{
        \pois(\zz; \dec{\rr})
    }
    =
    \dec{\rr} \cdot f(\enc{\dr}),
\end{equation}
where $\odot$ and $\cdot$ denote the Hadamard (element-wise) and vector products, respectively.

\subsection{Connection to sparse coding}

\Cref{eq:supp-nelbo-poisson} mirrors sparse coding due to the presence of the firing rate in the objective function. Furthermore, it follows the principle of predictive coding by design. Thus, our Poisson formulation of VAEs effectively unifies these two major themes in theoretical neuroscience. Let's explore this curious connection to sparse coding more closely below.

\subsection{Statistically independent neurons}

Suppose our \pvae has $K$ statistically independent neurons, and $\zz \in \ZZ_{\geq 0}^K$ is the spike count variable, where $\ZZ_{\geq 0} = \{0, 1, 2, \ldots\}$ is the set of non-negative integers. Let us use bold font $\dec{\rr}$ and $\enc{\dr}$ to refer to the firing rate vectors of the \dec{representation} and \enc{error} units, respectively. Recall that we allowed these variables to interact in a multiplicative way to construct the posterior rates, $\decenc{\lambda}_i(\xx) = \dec{r}_i \enc{\delta r}_i(\xx)$. More explicitly, we have:

\begin{equation}
\begin{aligned}
    q(\zz \vert \xx)
    &=
    \pois(
        \zz; \dec{\rr}\odot\enc{\dr}
    )
    =
    \prod_{i=1}^K\pois(
        z_i; \dec{r}_i\enc{\delta r}_i
    )
    = 
    \prod_{i=1}^K\frac{
        \decenc{\lambda}_i^{z_i}e^{-\decenc{\lambda}_i}
    }{z_i!},
    \\
    p(\zz)
    &=
    \pois(
        \zz; \dec{\rr}
    )
    =
    \prod_{i=1}^K\pois(
        z_i; \dec{r}_i
    )
    =
    \prod_{i=1}^K\frac{
        \dec{r}_i^{z_i}e^{-\dec{r}_i}
    }{z_i!}.
\end{aligned}
\end{equation}


Note that, unlike a standard Gaussian VAE, the prior in \pvae is parameterized using $\dec{\rr}$, which is learned from data along with the other parameters. Similar to standard Gaussian VAEs, $\enc{\dr}(\xx)$ is parameterized as a neural network.

\subsection{Linear decoder}

Following the sparse coding literature \cite{olshausen1996emergence}, we will now assume our decoder generates the input image $\xx \in \RR^M$ as a linear sum of $K$ basis elements, $\dec{\bm{\Phi}} \in \RR^{M \times K}$. Additionally, we choose a diagonal Gaussian distribution with fixed variance as our conditional likelihood, resulting in a mean squared error between the input $\xx$, and its reconstruction $\dec{\bm{\Phi}}\zz$.

Given these assumptions, the reconstruction loss for a VAE with approximate posterior $\enc{q}$ can be expressed as follows:
\begin{equation}\label{eq:supp-recon-linear}
    \LL_\mathrm{recon.}\left(
        \xx;
        \enc{q}
    \right) =
    \expect{
        \zz
        \,\sim\,
        \enc{q}(Z \vert X = \xx)
    }{
        \norm{\xx - \dec{\bm{\Phi}}\zz}_2^2
    }.
\end{equation}

For a linear decoder, the reconstruction term $\norm{\xx - \dec{\bm{\Phi}}\zz}_2^2$ contains only the first and second moments of $\zz$. Consequently, the expectation in \cref{eq:supp-recon-linear} can be analytically computed. This results in a close-form expression for the reconstruction loss, and consequently, its gradients as well.

In general, whenever the VAE decoder is linear, the following result holds:

\begin{empheq}[box=\fbox]{equation}\label{eq:supp-exact-recon-general}
\begin{aligned}
\mathcal{L}_\text{recon.}\left(
    \xx; \enc{q}, \dec{\bm{\Phi}}
\right)
=
\norm{\xx - \dec{\bm{\Phi}}
\expec{\enc{q}}{Z}}_2^2
+
\mathrm{Var}_\enc{q}[Z]^T
\mathrm{diag}(\dec{\bm{\Phi}}^T\dec{\bm{\Phi}}).
\end{aligned}
\end{empheq}

Note that a linear decoder is the only assumption we needed to obtain this closed-form solution. There are no restrictions on the form of the encoder: it can be linear, or as complicated as we want. We only have to compute the mean and variance of the posterior.

Specifically, for the Poisson case, we only need to know the following expectation values:

\begin{equation}\label{eq:supp-expectations}
\begin{aligned}
    \expect{
        \zz \sim \pois(\zz; \bm{\lambda})
    }{z_i}
    &=
    \lambda_i,
    \\
    \expect{
        \zz \sim \pois(\zz; \bm{\lambda})
    }{z_iz_j}
    &=
    \lambda_i\lambda_j + \delta_{ij}\lambda_i.
\end{aligned}
\end{equation} 

Here are the reconstruction losses for both Poisson and Gaussian VAEs with linear decoders, put side-by-side for comparison:

\begin{empheq}[box=\fbox]{equation}\label{eq:supp-exact-recon-both}
\begin{aligned}
    \text{Poisson:} \hspace{12.5mm} \mathcal{L}_\text{recon.}\left(
        \xx;
        \decenc{\bm{\lambda}},
        \dec{\bm{\Phi}}
    \right)
    &=
    \norm{\xx - \dec{\bm{\Phi}}\decenc{\bm{\lambda}}}_2^2
    +
    \decenc{\bm{\lambda}}^T\mathrm{diag}(\dec{\bm{\Phi}}^T\dec{\bm{\Phi}}),
    \\
    \text{Gaussian:} \quad\quad \mathcal{L}_\text{recon.}\left(
        \xx;
        \enc{\bmu},
        \enc{\bsig},
        \dec{\bm{\Phi}}
    \right)
    &=
    \norm{\xx - \dec{\bm{\Phi}}\enc{\bmu}}_2^2
    +
    (\enc{\bsig}^2)^T\mathrm{diag}(\dec{\bm{\Phi}}^T\dec{\bm{\Phi}}).
\end{aligned}
\end{empheq}

Given these assumptions, the NELBO (\cref{eq:supp-nelbo}) for \pvae with a linear decoder becomes:

\begin{empheq}[box=\fbox]{equation}\label{eq:supp-nelbo-poisson-linear}
    \LL_\text{SC-PVAE}\left(
        \xx;
        \enc{\dr},
        \dec{\rr},
        \dec{\bm{\Phi}}
    \right) =
    \norm{\xx - \dec{\bm{\Phi}}\decenc{\bm{\lambda}}}_2^2
    +
    \decenc{\bm{\lambda}}^T\mathrm{diag}(\dec{\bm{\Phi}}^T\dec{\bm{\Phi}})
    +
    \beta \sum_{i=1}^K
        \dec{r_i}
        f(\enc{\delta r_i}).
\end{empheq}

Recall that we have $f(y) = 1 - y + y\log y$ (see \cref{fig:poisson-residual}). We introduced the $\beta$ term here to control the trade-off between the reconstruction and the $\mathtt{KL}$ term \cite{higgins2017beta}. Additionally, we dropped the explicit dependence of $\enc{\dr}(\xx)$ on the input image $\xx$ to enhance readability.

\subsection{Linear encoder}

We can further simplify the \pvae architecture by making the encoder also linear. Let $\enc{\bm{W}} \in \RR^{K \times M}$ denote the encoder’s weight matrix, and assume an exponential link function mapping the input to residual firing rates, i.e., $\enc{\dr} = \exp(\enc{\bm{W}}\xx)$.

Starting from \cref{eq:supp-nelbo-poisson-linear}, substituting $\log \enc{\dr} = \enc{\bm{W}}\xx$, and rearranging terms yields the following loss function for the \linlin \pvae:
\begin{equation}\label{eq:supp-lindec-linenc-nelbo}
    \LL_\text{Lin-PVAE}
    = \decenc{\bm{\lambda}}^T\dec{\bm{\Phi}}^T\dec{\bm{\Phi}}\decenc{\bm{\lambda}}
    +
    \decenc{\bm{\lambda}}^T\mathrm{diag}(
        \dec{\bm{\Phi}}^T\dec{\bm{\Phi}}
        - \beta \bm{I}
    )
    +
    \decenc{\bm{\lambda}}^T(
        \beta \enc{\bm{W}} - 2 \dec{\bm{\Phi}}^T
    )\xx
    +
    \beta \sum_{i=1}^K \dec{r_i}
    +
    \xx^T\xx.
\end{equation}

\section{Architecture, training, and hyperparameter details}\label{sec:supp-architecture-training}

\subsection{Datasets: additional details}
We consider three datasets in this paper. We tile up the van Hateren dataset of natural images \cite{van1998independent} and CIFAR10 into $16 \times 16$ patches and apply whitening and contrast normalization using the code made available by \textcite{boutin2021sparse}. This operation results in the following total number of samples:

\begin{itemize}
    \item \textbf{van Hateren}: \quad $\# \text{train} = 107{,}520$, \quad $\# \text{validation} = 28{,}224$, 
    \item \textbf{\cifar}: \quad $\# \text{train} = 200{,}000$, \quad $\# \text{validation} = 40{,}000$.
\end{itemize}

We use the MNIST dataset primarily for the downstream classification task. After the training is done, we use the following train/validation split to evaluate the models:

\begin{itemize}
    \item \textbf{K-nearest neighbor classification} (\cref{tab:geom,tab:geom-full}): For this task, we only make use of the validation set for both training and testing of the classifier. We divide up the $N=10{,}000$ validation samples into two disjoint sets of $N=5{,}000$ samples each. We then draw random samples (without replacement) from the first half and use them for training the KNN classifier. We then test the performance on the other half.
    \item \textbf{Shattering dimensionality} (\cref{tab:geom,tab:geom-full}, last column): We use the entire MNIST training set ($N=60{,}000$ samples) to train logistic regression classifiers on extracted representations. We then test the results using the entire validation set ($N=10{,}000$ samples).
\end{itemize}

\subsection{Architecture details}
For sparse coding results, we focused on models with linear decoders. For the fully linear models (\cref{fig:phi,fig:dead-kl}) both the encoder and decoder were linear layers, without bias.

For the convolutional components, we use residual layers without batch norm. For van Hateren and \cifar datasets, the encoders had $5$ layers ($2 \times \mathrm{conv}$ each). The decoders had $8$ convolutional layers ($1 \times \mathrm{conv}$ each). For the MNIST dataset, the encoders had $7$ layers ($2 \times \mathrm{conv}$ each). The decoders had $10$ convolutional layers ($1 \times \mathrm{conv}$ each). For all convolutional encoders, the output from ResNet was followed by a learned pooling layer. The pooled output was then fed into a feed-forward layer inspired by Transformers \cite{vaswani2017attention}, which includes a layer norm as the final operation, the output of which was fed into a linear layer that projects features into posterior distribution parameters. For all convolutional decoders, nearest neighbor upsampling was performed to scale up the spatial dimension of reconstructions, as suggested by \textcite{child2021vdvae}.

We experimented with both leaky\_relu and swish activation functions \cite{ramachandran2018searching, elfwing2018sigmoid}, and found that swish consistently outperformed leaky\_relu in all our experiments across datasets and VAE models.

Please see our code for the full architecture details.

\subsection{Training details}

We used a variety of learning rates and batch sizes, depending on the dataset and architecture. For \linlin and \convlin models, we used $lr = 0.005$, and for \convconv models we used $lr = 0.002$. All models were trained using the AdaMax optimizer \cite{kingma2014adam} with a cosine learning rate schedule \cite{loshchilov2017sgdr}. Please see our code for the full details of training hyperparameters. Overall, we trained $195$ VAE models, $n=5$ seeds each, resulting in a total of $195 \times 5 = 975$ VAEs. For sparse coding models, we ran ISTA \cite{daubechies2004iterative, fista2009} and LCA \cite{rozell2008sparse} with 270 hyperparameter combinations each. Training all models took roughly a week on $8$ RTX 6000 Ada GPUs.

\paragraph{Temperature annealing for discrete VAEs.} We also annealed the temperature from a large value to a smaller value during the same first half of training for \pvae and \cvae. We found that the specific functional form of temperature annealing (e.g., linear, exponential, etc.) did not matter as much as the final temperature (\cref{fig:hard-fwd}). For both \pvae and \cvae, we start from $T_\text{start}=1.0$ and anneal down to $T_\text{stop} = 0.05$ for \pvae, and $T_\text{stop}=0.1$ for \cvae. We found that the \cvae performance was not very sensitive to the choice of $T_\text{stop}$, corroborating previous reports \cite{jang2017categorical, maddison2017concrete}.

The \pvae was relatively more sensitive to the value of $T_\text{stop}$, and we found marginal improvements when reducing it from $0.1$ to $0.05$. See \cref{fig:hard-fwd} for comprehensive experiments exploring the effect of the final temperature, as well as a \say{hard-forward} training method where we set $T=0$ in the forward pass (ensuring integer samples) and use a non-zero $T$ only during the backward pass (surrogate gradients). We find that our \say{relaxed Poisson} approach (\cref{fig:relaxed_poisson}) consistently outperforms the hard-forward approach.

\paragraph{KL annealing for VAEs.} For all VAE models, we annealed the $\mathtt{KL}$ term during the first half of the training, which is known to be an effective trick in training VAEs \cite{vahdat2020nvae, vafaii2023hierarchical, sonderby2016ladder, bowman2016generating, fu2019cyclical}.

\subsubsection{Training: sparse coding models}

To fit LCA and ISTA models, we explored a combination of 6 $\beta$ schedules (same $\beta$ as in \cref{eq:sparse-coding}), 3 numbers of iteration (for inference), 3 learning rates, and 5 different seeds (for dictionary initialization). The code for LCA was obtained from the public python library \say{\href{https://github.com/lanl/lca-pytorch}{lca-pytorch}} (\cite{ostiLCA}), and the code for ISTA was obtained from public \say{\href{https://github.com/rctn/sparsecoding/}{sparsecoding}} repository of the Redwood Center for Theoretical Neuroscience (with added clipping of coefficients to be nonnegative, following the thresholding step).

We explored learning rates of $\num{1e-1}$, $\num{1e-2}$, and $\num{1e-3}$. We trained all models for $100$ epochs. We scheduled the $\beta$ parameters linearly, starting from $\beta_\text{start}$, and stepped it up every five epochs by $\beta_\text{step}$, until it reached $\beta_\text{end}$. We explored the following $\beta$ schedules (expressed as $\beta_\text{start}$:$\beta_\text{end}$:$\beta_\text{step}$):

$$
0.05\text{:}0.7\text{:}0.1,\quad
0.01\text{:}0.1\text{:}0.01,\quad
0.1\text{:}1.0\text{:}0.1,\quad
0.05\text{:}0.7\text{:}0.05,\quad
0.05\text{:}0.5\text{:}0.05,\quad
0.1\text{:}0.1\text{:}0
$$

We also explored the inference iteration limits of $100$, $500$, and $900$ iterations. We selected the best fits to include in the main results shown in \cref{fig:phi,fig:rate-dist}.

\section{Supplementary results}\label{sec:supp-results}

In this section, we include additional results that further support those reported in the main paper, including:

\Cref{tab:nelbo-full} contains the negative ELBO values for all VAE models with a linear decoder. This table reveals a comparable performance between using Monte Carlo samples to estimate gradients, versus optimizing the exact loss (see \cref{eq:sc-pvae-nelbo,eq:supp-exact-recon-general,eq:supp-exact-recon-both,eq:supp-nelbo-poisson-linear}), highlighting the effectiveness of our Poisson reparameterization algorithm.

\begin{figure}[ht!]
    \centering
    \includegraphics[width=\linewidth]{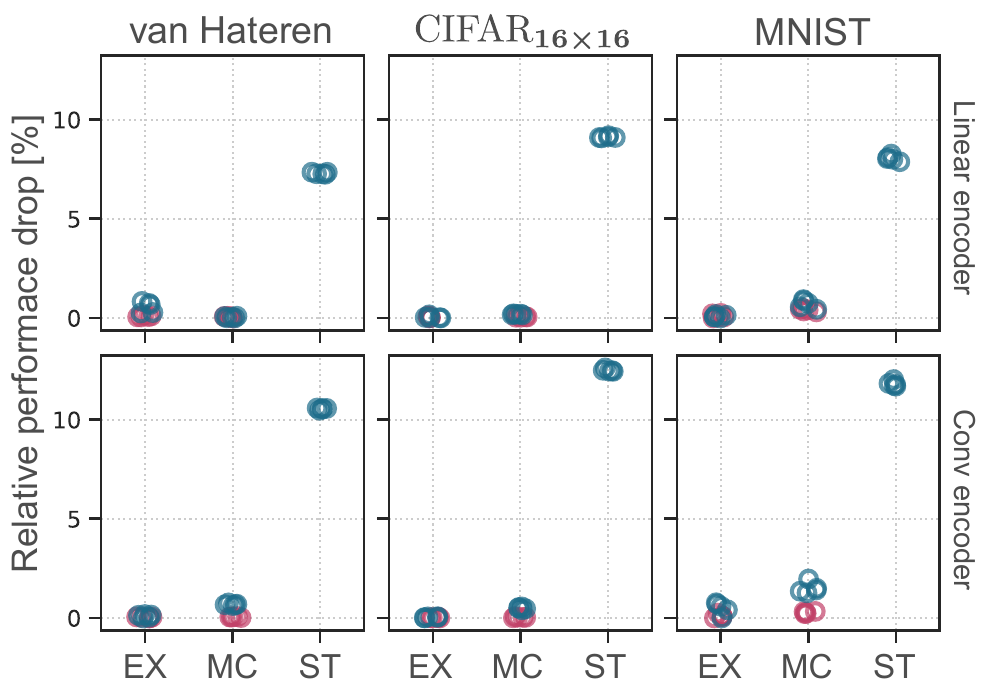}
    \caption{
        Performance drop relative to the best fit. Blue circles indicate \pvae results, red circles indicate \gvae results, and each set of $n=5$ circles corresponds to five random initializations. Using Monte Carlo samples \cite{mohamed2020monte} and our Poisson reparameterization trick (\Cref{algo:rsample}) to estimate gradients performs comparably to using exact gradients (see \cref{eq:supp-exact-recon-general,eq:supp-exact-recon-both,eq:supp-nelbo-poisson-linear}). \Cref{tab:exmcst} provides a tabular summary of these results.
        EX, exact, MC, Monte-Carlo, ST, straight-through \cite{bengio2013estimating}.
    }
    \label{fig:exmcst}
\end{figure}

\Cref{fig:exmcst} uses the same data from the main paper \Cref{tab:exmcst} to visualize the effects.

\begin{figure}[ht!]
    \centering
    \includegraphics[width=0.65\linewidth]{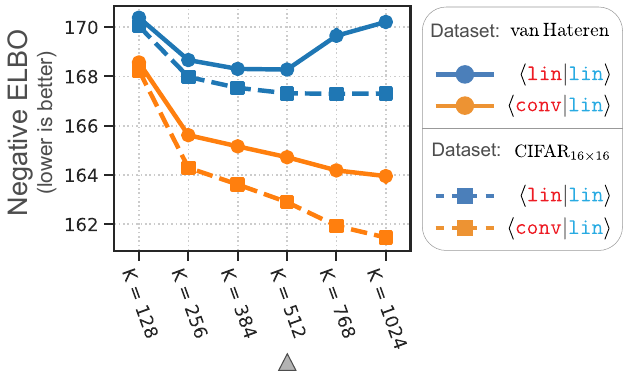}
    \caption{
        The effect of latent dimensionality on model performance across datasets and encoder architectures. For all convolutional encoder cases, ELBO improves as a function of latent dimensionality. However, for linear encoders, we see that the van Hateren dataset starts to overfit for $K > 512$, and it stagnates for the \cifar dataset. In conclusion, more expressive encoders can find nonlinear features, represented using additional latent dimensions, but simple linear encoders struggle to utilize additional dimensions. The gray triangle indicates the setting used in the main results.
    }
    \label{fig:dim-effect}
\end{figure}

\Cref{fig:dim-effect} shows the dependence of loss on latent dimensionality. We find that increasing the number of latent dimensions consistently improves ELBO for \convlin architectures, but \linlin models either overfit (for van Hateren) or fail to improve (for \cifar) once $K$ becomes large.

\begin{figure}[ht!]
    \centering
    \includegraphics[width=0.85\linewidth]{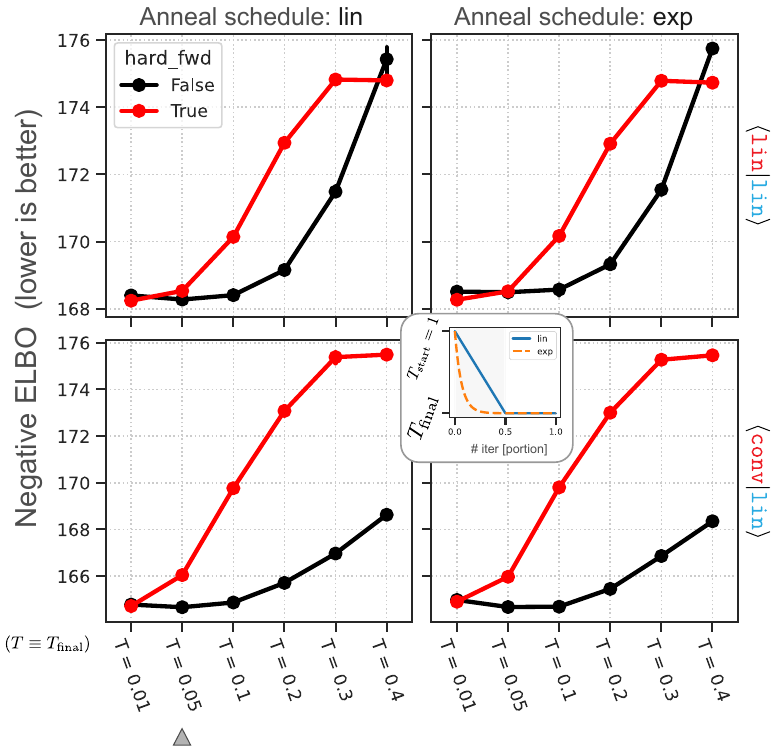}
    \caption{
        Performance as a function of the final temperature ($T_\text{final}$), annealing schedule (linear vs.~exponential; inset), and the ``hard-forward'' approach. The hard-forward approach uses exact integer samples ($T=0$) in the forward pass and applies nonzero temperatures only in the backward pass (i.e., ``surrogate gradients''). Although all results are evaluated at $T=0$ during testing, the hard-forward approach still underperforms our ``relaxed Poisson'' method (\cref{fig:relaxed_poisson}), which employs continuous (floating) samples during training due to a non-zero $T$ (\Cref{algo:rsample}). The gray triangle indicates the setting used in the main results: $T_\text{final} = 0.05$ with a linear annealing schedule.
    }
    \label{fig:hard-fwd}
\end{figure}

\Cref{fig:hard-fwd} demonstrates the robustness of our Poisson reparameterization trick (\Cref{algo:rsample}) to variations in the temperature parameter. Importantly, we also explore a \say{hard-forward} training approach, where we fix $T=0$ during the forward pass but allow $T > 0$ in the backward pass. This is also known as \textit{surrogate gradients}. We find that, somewhat surprisingly, this hard-forward method performs significantly worse than our \say{relaxed Poisson} approach (\cref{fig:relaxed_poisson}).

\begin{figure}[ht!]
    \centering
    \includegraphics[width=\linewidth]{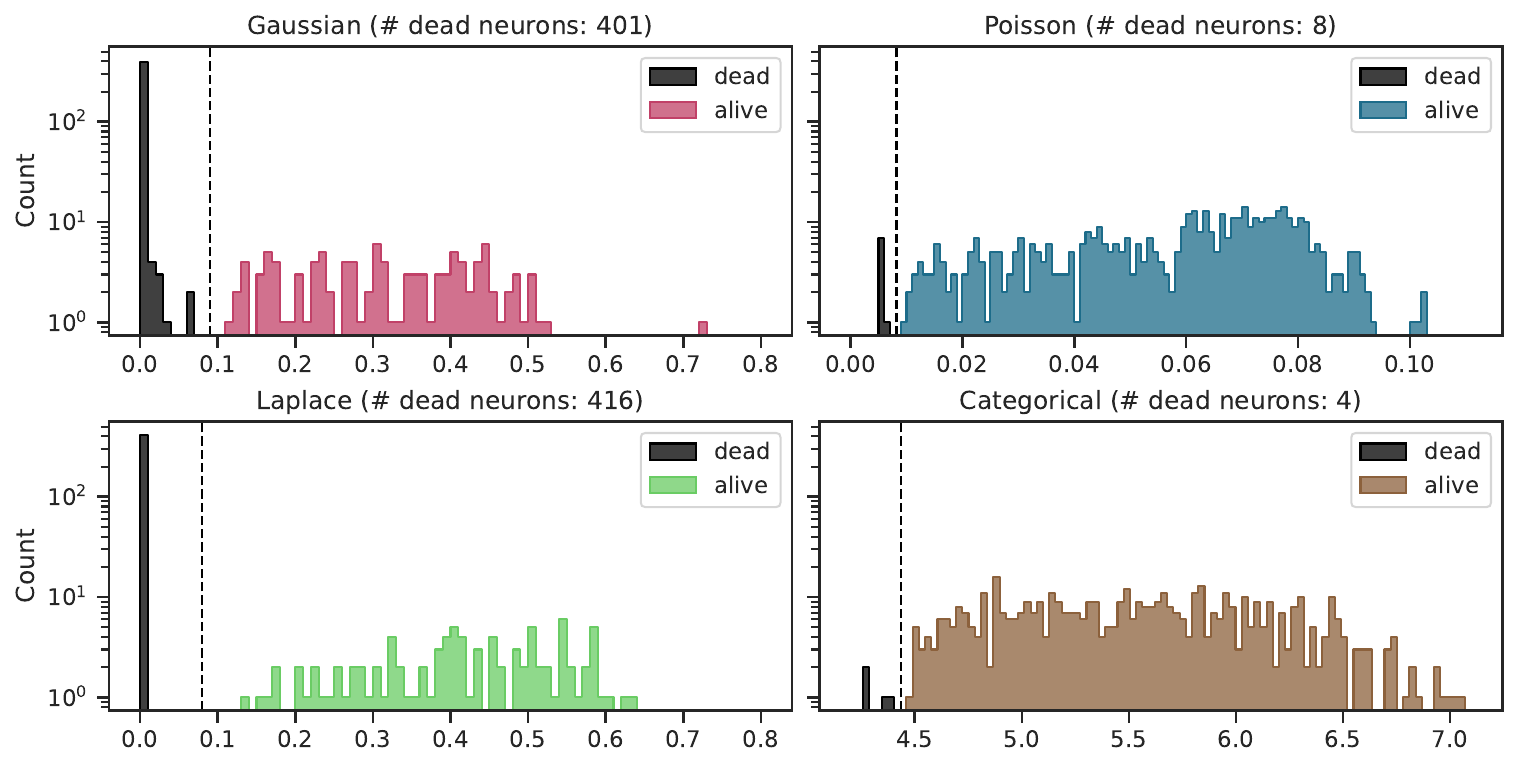}
    \caption{
        Identifying dead neurons using a histogram-based method. We bin the $\mathtt{KL}$ values and determine the gap between small values and larger ones. We identify neurons with $\mathtt{KL}$ values lower than the identified threshold (black dashed lines) and pronounce them dead. The figure shows the distribution of $\mathtt{KL}$ values over all neurons ($K=512$) for \pvae, \gvae, and \lvae. The $\mathtt{KL}$ term is a single number for the \cvae because its latent space consists of a single one-hot categorical distribution with $K=512$ categories. Therefore, for the \cvae, we use the distribution of decoder weight norms instead. These are the same models shown in \cref{fig:phi}, where both encoder and decoder are linear. \Cref{tab:active} uses this method to quantify the proportion of active neurons for VAEs across different datasets and the choice of encoder architectures.
    }
    \label{fig:dead-kl}
\end{figure}

\Cref{fig:dead-kl} shows how the distribution of $\mathtt{KL}$ values (or the norm of decoder weights in the case of linear decoders) can be used to determine dead neurons that don't contribute to the encoding of information.

\Cref{tab:geom-full} contains the full set of downstream classification results. Related to \Cref{tab:geom}.

\begin{figure}[ht!]
    \centering
    \includegraphics[width=0.7\linewidth]{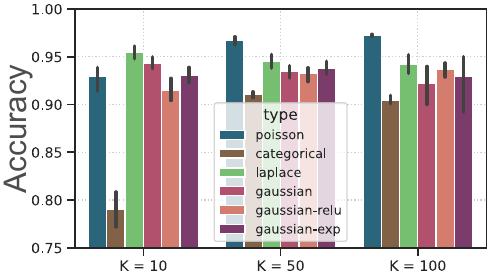}
    \caption{
        Downstream classification performance using a simple linear classifier. After unsupervised training of \convconv VAEs on MNIST, we extracted latent representations and applied logistic regression. For $K=100$, \pvae achieves the highest accuracy, while for $K=10$, both \lvae and \gvae outperform it.
    }
    \label{fig:logreg}
\end{figure}

\Cref{fig:logreg} shows the performance of a simple linear classifier (logistic regression) trained on unsupervised representations learned by various \convconv VAEs. We find that increasing the latent dimension ($K$) generally improves the performance of \pvae, but at lower dimensions, other methods like \lvae and \gvae can outperform it.

\begin{figure}[ht!]
    \centering
    \includegraphics[width=\linewidth]{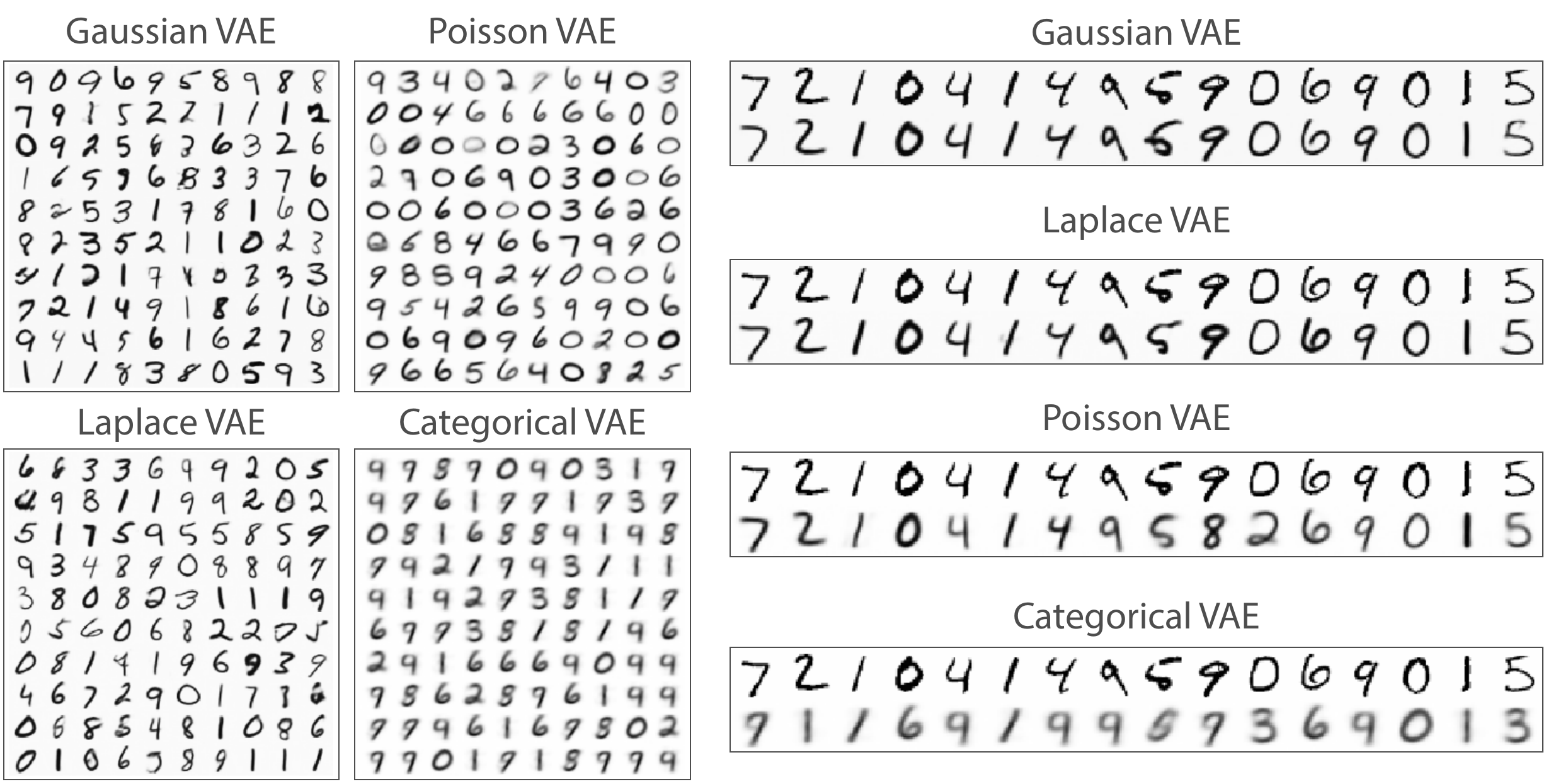}
    \caption{
        Generated samples (left) and reconstruction performance (right). These results shown here are from models with a \convconv architectures and latent dimensionality of $K=10$.
    }
    \label{fig:recon+sample}
\end{figure}

\Cref{fig:recon+sample} shows MNIST samples generated from the latent space of different \convconv VAE models, as well as their reconstruction performance.

\begin{table}[ht!]
    \caption{The reparameterized gradient estimators work as well as exact ones, across datasets and \enc{encoder} architectures (linear vs.~conv). Note that exact gradients are only computable for linear \dec{decoders} (see \cref{eq:supp-exact-recon-general,eq:supp-exact-recon-both,eq:supp-nelbo-poisson-linear}). The values are negative ELBO (lower is better), shown as \entry{\mathrm{mean}}{99\%} confidence interval calculated from $n=5$ different random initializations. For MNIST, our use of Gaussian conditional likelihoods means the numerical performance values are not directly comparable to studies that use binarized MNIST with a cross-entropy decoder. EX, exact, MC, Monte-Carlo, ST, straight-through \cite{bengio2013estimating}. See also \Cref{tab:exmcst} and supplementary \cref{fig:exmcst}.}
    \label{tab:nelbo-full}
    \centering
    \begin{tabular}{l@{\hspace{1mm}}lccc}
        \toprule
        Model
        &
        &
        \begin{tabularx}{34mm}{CC}
            \multicolumn{2}{c}{van Hateren}\\
            \midrule
            {\small\linlin} & {\small\convlin}
        \end{tabularx}
        &
        \begin{tabularx}{34mm}{CC}
            \multicolumn{2}{c}{\cifar}\\
            \midrule
            {\small\linlin} & {\small\convlin}
        \end{tabularx}
        &
        \begin{tabularx}{34mm}{CC}
            \multicolumn{2}{c}{MNIST}\\
            \midrule
            {\small\linlin} & {\small\convlin}
        \end{tabularx}
        \\
        \midrule
        \pvae
        &
        \begin{tabularx}{5mm}{C}
            EX\\MC\\ST
        \end{tabularx}
        &
        \begin{tabularx}{34mm}{CC}
            \entry{168.0}{.8} & \entry{162.4}{.2} \\ 
            \entry{167.2}{.1} & \entry{163.4}{.1} \\ 
            \entry{179.3}{.1} & \entry{179.4}{.1}
        \end{tabularx}
        &
        \begin{tabularx}{34mm}{CC}
            \entry{167.1}{.2} & \entry{162.1}{.1} \\ 
            \entry{167.3}{.1} & \entry{162.9}{.2} \\ 
            \entry{182.3}{.1} & \entry{182.3}{.2}
        \end{tabularx}
        &
        \begin{tabularx}{34mm}{CC}
            \entry{41.5}{.1} & \entry{39.7}{.2} \\ 
            \entry{41.7}{.2} & \entry{40.1}{.2} \\ 
            \entry{44.8}{.1} & \entry{44.2}{.1}
        \end{tabularx}
        \\
        \midrule
        \gvae
        &
        \begin{tabularx}{5mm}{C}
            EX\\MC
        \end{tabularx}
        &
        \begin{tabularx}{34mm}{CC}
            \entry{160.3}{.1} & \entry{154.4}{.1} \\ 
            \entry{160.3}{.1} & \entry{154.4}{.1}
        \end{tabularx}
        &
        \begin{tabularx}{34mm}{CC}
            \entry{165.9}{.1} & \entry{149.2}{.0} \\ 
            \entry{165.9}{.1} & \entry{149.2}{.1}
        \end{tabularx}
        &
        \begin{tabularx}{34mm}{CC}
            \entry{40.6}{.1} & \entry{40.0}{.1} \\ 
            \entry{40.7}{.1} & \entry{40.1}{.0}
        \end{tabularx}
        \\
        \midrule
        \cvae
        &
        \begin{tabularx}{5mm}{C}
            EX\\MC\\ST
        \end{tabularx}
        &
        \begin{tabularx}{34mm}{CC}
            \entry{174.9}{.1} & \entry{186.3}{.8} \\ 
            \entry{170.5}{.1} & \entry{171.9}{.2} \\ 
            \entry{174.2}{.2} & \entry{181.1}{.3}
        \end{tabularx}
        &
        \begin{tabularx}{34mm}{CC}
            \entry{177.1}{.1} & \entry{180.6}{.5} \\ 
            \entry{174.7}{.1} & \entry{176.5}{.1} \\ 
            \entry{180.2}{.0} & \entry{185.6}{.2}
        \end{tabularx}
        &
        \begin{tabularx}{34mm}{CC}
            \entry{56.1}{.7} & \entry{59.1}{.0} \\ 
            \entry{39.7}{.2} & \entry{59.1}{.0} \\ 
            \entry{49.3}{.1} & \entry{63.8}{3.4}
        \end{tabularx}
        \\
        \midrule
        \lvae
        &
        \begin{tabularx}{5mm}{C}
            EX\\MC\
        \end{tabularx}
        &
        \begin{tabularx}{34mm}{CC}
            \entry{167.3}{.0} & \entry{159.0}{.2} \\ 
            \entry{167.3}{.0} & \entry{159.2}{.2}
        \end{tabularx}
        &
        \begin{tabularx}{34mm}{CC}
            \entry{170.1}{.1} & \entry{154.3}{.1} \\ 
            \entry{170.1}{.1} & \entry{154.5}{.1}
        \end{tabularx}
        &
        \begin{tabularx}{34mm}{CC}
            \entry{42.1}{.1} & \entry{41.0}{.0} \\ 
            \entry{42.1}{.0} & \entry{41.0}{.0}
        \end{tabularx}
        \\
        \bottomrule
  \end{tabular}
\end{table}

\begin{table}[ht!]
    \caption{Geometry of representations. Full set of results. Related to \Cref{tab:geom}.}
    \label{tab:geom-full}
    \centering
    \begin{tabular}{c l c c}
        \toprule
        \begin{tabular}{c}
            Latent\\dim.
        \end{tabular}
        &
        \hspace{2.5mm}Model
        &
        \begin{tabularx}{65mm}{CCC}
            \multicolumn{3}{c}{KNN classification ($N,$ \# labeled samples)}\\
            \midrule
            $N=200$ & $N=1{,}000$ & $N=5{,}000$
        \end{tabularx}
        &
        \begin{tabular}{c}
            Shattering\\dim.
        \end{tabular}
        \\
        \midrule
        $K = 10$
        &
        \begin{tabular}{l}
            \pvae \\
            \cvae \\
            \lvae \\
            \gvae \\
            \grelu \\
            \gexp
        \end{tabular}
        &
        \begin{tabularx}{65mm}{CCC}
            \entry{\bf0.815}{.002} & \entry{\bf0.919}{.001} & \entry{\bf0.946}{.017} \\ 
            \entry{0.705}{.002} & \entry{0.800}{.002} & \entry{0.853}{.040} \\ 
            \entry{0.757}{.003} & \entry{0.869}{.002} & \entry{\bf0.924}{.028} \\ 
            \entry{0.673}{.003} & \entry{0.813}{.002} & \entry{0.891}{.033} \\ 
            \entry{0.694}{.003} & \entry{0.817}{.003} & \entry{0.877}{.045} \\ 
            \entry{0.642}{.003} & \entry{0.784}{.002} & \entry{0.863}{.032}
        \end{tabularx}
        &
        \begin{tabular}{c}
            \entry{\bf0.797}{.009} \\ 
            \entry{\bf0.795}{.006} \\ 
            \entry{0.751}{.008} \\ 
            \entry{0.758}{.007} \\ 
            \entry{0.762}{.007} \\ 
            \entry{0.737}{.008}
        \end{tabular}
        \\
        \midrule
        $K = 50$
        &
        \begin{tabular}{l}
            \pvae \\
            \cvae \\
            \lvae \\
            \gvae \\
            \grelu \\
            \gexp
        \end{tabular}
        &
        \begin{tabularx}{65mm}{CCC}
            \entry{\bf0.825}{.002} & \entry{\bf0.927}{.001} & \entry{\bf0.957}{.005} \\ 
            \entry{0.770}{.002} & \entry{0.880}{.001} & \entry{0.920}{.009} \\ 
            \entry{0.710}{.003} & \entry{0.836}{.003} & \entry{0.902}{.038} \\ 
            \entry{0.604}{.003} & \entry{0.746}{.002} & \entry{0.837}{.022} \\ 
            \entry{0.710}{.002} & \entry{0.844}{.002} & \entry{0.904}{.026} \\ 
            \entry{0.694}{.003} & \entry{0.836}{.002} & \entry{0.906}{.027}
        \end{tabularx}
        &
        \begin{tabular}{c}
            \entry{\bf0.935}{.003} \\ 
            \entry{0.899}{.004} \\ 
            \entry{0.770}{.007} \\ 
            \entry{0.743}{.007} \\ 
            \entry{0.786}{.006} \\ 
            \entry{0.762}{.007}
        \end{tabular}
        \\
        \midrule
        $K = 100$
        &
        \begin{tabular}{l}
            \pvae \\
            \cvae \\
            \lvae \\
            \gvae \\
            \grelu \\
            \gexp
        \end{tabular}
        &
        \begin{tabularx}{65mm}{CCC}
            \entry{\bf0.807}{.002} & \entry{\bf0.925}{.001} & \entry{\bf0.958}{.013} \\ 
            \entry{0.753}{.002} & \entry{0.876}{.001} & \entry{0.925}{.005} \\ 
            \entry{0.701}{.004} & \entry{0.830}{.003} & \entry{\bf0.896}{.046} \\ 
            \entry{0.636}{.003} & \entry{0.789}{.002} & \entry{0.875}{.024} \\ 
            \entry{0.757}{.002} & \entry{0.881}{.001} & \entry{\bf0.933}{.019} \\ 
            \entry{0.695}{.003} & \entry{0.846}{.002} & \entry{0.918}{.024}
        \end{tabularx}
        &
        \begin{tabular}{c}
            \entry{\bf0.949}{.002} \\ 
            \entry{0.884}{.004} \\ 
            \entry{0.767}{.007} \\ 
            \entry{0.763}{.007} \\ 
            \entry{0.818}{.006} \\ 
            \entry{0.793}{.006}
        \end{tabular}
        \\
        \bottomrule
  \end{tabular}
\end{table}

\clearpage
\section*{NeurIPS Paper Checklist}

\begin{enumerate}

\item {\bf Claims}
    \item[] Question: Do the main claims made in the abstract and introduction accurately reflect the paper's contributions and scope?
    \item[] Answer: \answerYes{} 
    \item[] Justification: we provide comprehensive theoretical and empirical evidence to support our claims of (1) introducing the \pvae and its reparameterization trick; (2) \pvae containing amortized sparse coding as a special case; (3) \pvae largely avoiding posterior collapse; and (4) \pvae facilitating linear separability of categories at better sample efficiency, in \cref{sec:theory,sec:experiments}, and supplemental \cref{sec:supp-architecture-training,sec:supp-results,sec:supp-derivations}.
    \item[] Guidelines:
    \begin{itemize}
        \item The answer NA means that the abstract and introduction do not include the claims made in the paper.
        \item The abstract and/or introduction should clearly state the claims made, including the contributions made in the paper and important assumptions and limitations. A No or NA answer to this question will not be perceived well by the reviewers. 
        \item The claims made should match theoretical and experimental results, and reflect how much the results can be expected to generalize to other settings. 
        \item It is fine to include aspirational goals as motivation as long as it is clear that these goals are not attained by the paper. 
    \end{itemize}

\item {\bf Limitations}
    \item[] Question: Does the paper discuss the limitations of the work performed by the authors?
    \item[] Answer: \answerYes{} 
    \item[] Justification: The limitations of (1) Poisson possibly not being a perfect description of cortical activity, and (2) amortization gap, are shown explicitly and thoroughly discussed in \cref{sec:experiments,sec:conclusion}. Specifically, we have a dedicated paragraph for limitations in \cref{sec:conclusion}. We evaluated our claims using multiple well-known datasets such as the van Hateren natural images \cite{van1998independent}, CIFAR10, and MNIST, on tasks such as reconstruction, sparse coding, and classification.
    \item[] Guidelines:
    \begin{itemize}
        \item The answer NA means that the paper has no limitation while the answer No means that the paper has limitations, but those are not discussed in the paper. 
        \item The authors are encouraged to create a separate "Limitations" section in their paper.
        \item The paper should point out any strong assumptions and how robust the results are to violations of these assumptions (e.g., independence assumptions, noiseless settings, model well-specification, asymptotic approximations only holding locally). The authors should reflect on how these assumptions might be violated in practice and what the implications would be.
        \item The authors should reflect on the scope of the claims made, e.g., if the approach was only tested on a few datasets or with a few runs. In general, empirical results often depend on implicit assumptions, which should be articulated.
        \item The authors should reflect on the factors that influence the performance of the approach. For example, a facial recognition algorithm may perform poorly when image resolution is low or images are taken in low lighting. Or a speech-to-text system might not be used reliably to provide closed captions for online lectures because it fails to handle technical jargon.
        \item The authors should discuss the computational efficiency of the proposed algorithms and how they scale with dataset size.
        \item If applicable, the authors should discuss possible limitations of their approach to address problems of privacy and fairness.
        \item While the authors might fear that complete honesty about limitations might be used by reviewers as grounds for rejection, a worse outcome might be that reviewers discover limitations that aren't acknowledged in the paper. The authors should use their best judgment and recognize that individual actions in favor of transparency play an important role in developing norms that preserve the integrity of the community. Reviewers will be specifically instructed to not penalize honesty concerning limitations.
    \end{itemize}

\item {\bf Theory Assumptions and Proofs}
    \item[] Question: For each theoretical result, does the paper provide the full set of assumptions and a complete (and correct) proof?
    \item[] Answer: \answerYes{} 
    \item[] Justification: We provide the full derivation of the \pvae loss function, which is self-contained in the paper (\cref{sec:theory}) and supplement (\cref{sec:supp-derivations}).
    \item[] Guidelines:
    \begin{itemize}
        \item The answer NA means that the paper does not include theoretical results. 
        \item All the theorems, formulas, and proofs in the paper should be numbered and cross-referenced.
        \item All assumptions should be clearly stated or referenced in the statement of any theorems.
        \item The proofs can either appear in the main paper or the supplemental material, but if they appear in the supplemental material, the authors are encouraged to provide a short proof sketch to provide intuition. 
        \item Inversely, any informal proof provided in the core of the paper should be complemented by formal proofs provided in appendix or supplemental material.
        \item Theorems and Lemmas that the proof relies upon should be properly referenced. 
    \end{itemize}

    \item {\bf Experimental Result Reproducibility}
    \item[] Question: Does the paper fully disclose all the information needed to reproduce the main experimental results of the paper to the extent that it affects the main claims and/or conclusions of the paper (regardless of whether the code and data are provided or not)?
    \item[] Answer: \answerYes{} 
    \item[] Justification: We disclose all details relating to the algorithm, including the optimization objective (\cref{eq:pvae-nelbo}), architecture and training details (\cref{sec:supp-architecture-training}), and pseudo-code for Poisson reparameterized sampling (\Cref{algo:rsample}). In addition, we intend to release all code and data needed for replicating our work. 
    \item[] Guidelines:
    \begin{itemize}
        \item The answer NA means that the paper does not include experiments.
        \item If the paper includes experiments, a No answer to this question will not be perceived well by the reviewers: Making the paper reproducible is important, regardless of whether the code and data are provided or not.
        \item If the contribution is a dataset and/or model, the authors should describe the steps taken to make their results reproducible or verifiable. 
        \item Depending on the contribution, reproducibility can be accomplished in various ways. For example, if the contribution is a novel architecture, describing the architecture fully might suffice, or if the contribution is a specific model and empirical evaluation, it may be necessary to either make it possible for others to replicate the model with the same dataset, or provide access to the model. In general. releasing code and data is often one good way to accomplish this, but reproducibility can also be provided via detailed instructions for how to replicate the results, access to a hosted model (e.g., in the case of a large language model), releasing of a model checkpoint, or other means that are appropriate to the research performed.
        \item While NeurIPS does not require releasing code, the conference does require all submissions to provide some reasonable avenue for reproducibility, which may depend on the nature of the contribution. For example
        \begin{enumerate}
            \item If the contribution is primarily a new algorithm, the paper should make it clear how to reproduce that algorithm.
            \item If the contribution is primarily a new model architecture, the paper should describe the architecture clearly and fully.
            \item If the contribution is a new model (e.g., a large language model), then there should either be a way to access this model for reproducing the results or a way to reproduce the model (e.g., with an open-source dataset or instructions for how to construct the dataset).
            \item We recognize that reproducibility may be tricky in some cases, in which case authors are welcome to describe the particular way they provide for reproducibility. In the case of closed-source models, it may be that access to the model is limited in some way (e.g., to registered users), but it should be possible for other researchers to have some path to reproducing or verifying the results.
        \end{enumerate}
    \end{itemize}

\item {\bf Open access to data and code}
    \item[] Question: Does the paper provide open access to the data and code, with sufficient instructions to faithfully reproduce the main experimental results, as described in supplemental material?
    \item[] Answer: \answerYes{} 
    \item[] Justification: Our code, data, and model checkpoints are available from the following GitHub repository: \href{https://github.com/hadivafaii/PoissonVAE}{\color{purple}https://github.com/hadivafaii/PoissonVAE}. 
    \item[] Guidelines:
    \begin{itemize}
        \item The answer NA means that paper does not include experiments requiring code.
        \item Please see the NeurIPS code and data submission guidelines (\url{https://nips.cc/public/guides/CodeSubmissionPolicy}) for more details.
        \item While we encourage the release of code and data, we understand that this might not be possible, so “No” is an acceptable answer. Papers cannot be rejected simply for not including code, unless this is central to the contribution (e.g., for a new open-source benchmark).
        \item The instructions should contain the exact command and environment needed to run to reproduce the results. See the NeurIPS code and data submission guidelines (\url{https://nips.cc/public/guides/CodeSubmissionPolicy}) for more details.
        \item The authors should provide instructions on data access and preparation, including how to access the raw data, preprocessed data, intermediate data, and generated data, etc.
        \item The authors should provide scripts to reproduce all experimental results for the new proposed method and baselines. If only a subset of experiments are reproducible, they should state which ones are omitted from the script and why.
        \item At submission time, to preserve anonymity, the authors should release anonymized versions (if applicable).
        \item Providing as much information as possible in supplemental material (appended to the paper) is recommended, but including URLs to data and code is permitted.
    \end{itemize}

\item {\bf Experimental Setting/Details}
    \item[] Question: Does the paper specify all the training and test details (e.g., data splits, hyperparameters, how they were chosen, type of optimizer, etc.) necessary to understand the results?
    \item[] Answer: \answerYes{} 
    \item[] Justification: All details about how the data was used for training and testing, as well as which hyperparameters were used, are available at \cref{sec:supp-architecture-training}. In addition, the provided code replicates our results and therefore contains all details of implementation.
    \item[] Guidelines:
    \begin{itemize}
        \item The answer NA means that the paper does not include experiments.
        \item The experimental setting should be presented in the core of the paper to a level of detail that is necessary to appreciate the results and make sense of them.
        \item The full details can be provided either with the code, in appendix, or as supplemental material.
    \end{itemize}

\item {\bf Experiment Statistical Significance}
    \item[] Question: Does the paper report error bars suitably and correctly defined or other appropriate information about the statistical significance of the experiments?
    \item[] Answer: \answerYes{} 
    \item[] Justification: As stated in \cref{sec:experiments}, the paper reports confidence intervals and $t$-test significance tests, using false discovery rate (FDR) correction for multiple comparisons. The exact implementation details are included in the provided code for reproducibility.
    \item[] Guidelines:
    \begin{itemize}
        \item The answer NA means that the paper does not include experiments.
        \item The authors should answer "Yes" if the results are accompanied by error bars, confidence intervals, or statistical significance tests, at least for the experiments that support the main claims of the paper.
        \item The factors of variability that the error bars are capturing should be clearly stated (for example, train/test split, initialization, random drawing of some parameter, or overall run with given experimental conditions).
        \item The method for calculating the error bars should be explained (closed form formula, call to a library function, bootstrap, etc.)
        \item The assumptions made should be given (e.g., Normally distributed errors).
        \item It should be clear whether the error bar is the standard deviation or the standard error of the mean.
        \item It is OK to report 1-sigma error bars, but one should state it. The authors should preferably report a 2-sigma error bar than state that they have a 96\% CI, if the hypothesis of Normality of errors is not verified.
        \item For asymmetric distributions, the authors should be careful not to show in tables or figures symmetric error bars that would yield results that are out of range (e.g. negative error rates).
        \item If error bars are reported in tables or plots, The authors should explain in the text how they were calculated and reference the corresponding figures or tables in the text.
    \end{itemize}

\item {\bf Experiments Compute Resources}
    \item[] Question: For each experiment, does the paper provide sufficient information on the computer resources (type of compute workers, memory, time of execution) needed to reproduce the experiments?
    \item[] Answer: \answerYes{} 
    \item[] Justification: We provide details about our compute resources (GPUs), and duration of training in section \ref{sec:experiments}.
    \item[] Guidelines:
    \begin{itemize}
        \item The answer NA means that the paper does not include experiments.
        \item The paper should indicate the type of compute workers CPU or GPU, internal cluster, or cloud provider, including relevant memory and storage.
        \item The paper should provide the amount of compute required for each of the individual experimental runs as well as estimate the total compute. 
        \item The paper should disclose whether the full research project required more compute than the experiments reported in the paper (e.g., preliminary or failed experiments that didn't make it into the paper). 
    \end{itemize}
    
\item {\bf Code Of Ethics}
    \item[] Question: Does the research conducted in the paper conform, in every respect, with the NeurIPS Code of Ethics \url{https://neurips.cc/public/EthicsGuidelines}?
    \item[] Answer: \answerYes{} 
    \item[] Justification: Our paper follows the code of ethics, including preserving anonymity (such as in releasing code anonymously).
    \item[] Guidelines:
    \begin{itemize}
        \item The answer NA means that the authors have not reviewed the NeurIPS Code of Ethics.
        \item If the authors answer No, they should explain the special circumstances that require a deviation from the Code of Ethics.
        \item The authors should make sure to preserve anonymity (e.g., if there is a special consideration due to laws or regulations in their jurisdiction).
    \end{itemize}

\item {\bf Broader Impacts}
    \item[] Question: Does the paper discuss both potential positive societal impacts and negative societal impacts of the work performed?
    \item[] Answer: \answerNA{} 
    \item[] Justification: Our paper is considered foundational research, and does not target practical tasks that can be deployed outside of the research field. Thus we do not anticipate negative social impacts from this work.
    \item[] Guidelines:
    \begin{itemize}
        \item The answer NA means that there is no societal impact of the work performed.
        \item If the authors answer NA or No, they should explain why their work has no societal impact or why the paper does not address societal impact.
        \item Examples of negative societal impacts include potential malicious or unintended uses (e.g., disinformation, generating fake profiles, surveillance), fairness considerations (e.g., deployment of technologies that could make decisions that unfairly impact specific groups), privacy considerations, and security considerations.
        \item The conference expects that many papers will be foundational research and not tied to particular applications, let alone deployments. However, if there is a direct path to any negative applications, the authors should point it out. For example, it is legitimate to point out that an improvement in the quality of generative models could be used to generate deepfakes for disinformation. On the other hand, it is not needed to point out that a generic algorithm for optimizing neural networks could enable people to train models that generate Deepfakes faster.
        \item The authors should consider possible harms that could arise when the technology is being used as intended and functioning correctly, harms that could arise when the technology is being used as intended but gives incorrect results, and harms following from (intentional or unintentional) misuse of the technology.
        \item If there are negative societal impacts, the authors could also discuss possible mitigation strategies (e.g., gated release of models, providing defenses in addition to attacks, mechanisms for monitoring misuse, mechanisms to monitor how a system learns from feedback over time, improving the efficiency and accessibility of ML).
    \end{itemize}
    
\item {\bf Safeguards}
    \item[] Question: Does the paper describe safeguards that have been put in place for responsible release of data or models that have a high risk for misuse (e.g., pretrained language models, image generators, or scraped datasets)?
    \item[] Answer: \answerNA{} 
    \item[] Justification: Our paper utilizes publicly domain datasets (not scraped), and poses no safety risks.
    \item[] Guidelines:
    \begin{itemize}
        \item The answer NA means that the paper poses no such risks.
        \item Released models that have a high risk for misuse or dual-use should be released with necessary safeguards to allow for controlled use of the model, for example by requiring that users adhere to usage guidelines or restrictions to access the model or implementing safety filters. 
        \item Datasets that have been scraped from the Internet could pose safety risks. The authors should describe how they avoided releasing unsafe images.
        \item We recognize that providing effective safeguards is challenging, and many papers do not require this, but we encourage authors to take this into account and make a best faith effort.
    \end{itemize}

\item {\bf Licenses for existing assets}
    \item[] Question: Are the creators or original owners of assets (e.g., code, data, models), used in the paper, properly credited and are the license and terms of use explicitly mentioned and properly respected?
    \item[] Answer: \answerYes{} 
    \item[] Justification: We properly cite papers that introduce algorithms (such as LCA), datasets (such as MNIST, CIFAR10, van Hateren), and code (such as LCA and ISTA).
    \item[] Guidelines:
    \begin{itemize}
        \item The answer NA means that the paper does not use existing assets.
        \item The authors should cite the original paper that produced the code package or dataset.
        \item The authors should state which version of the asset is used and, if possible, include a URL.
        \item The name of the license (e.g., CC-BY 4.0) should be included for each asset.
        \item For scraped data from a particular source (e.g., website), the copyright and terms of service of that source should be provided.
        \item If assets are released, the license, copyright information, and terms of use in the package should be provided. For popular datasets, \url{paperswithcode.com/datasets} has curated licenses for some datasets. Their licensing guide can help determine the license of a dataset.
        \item For existing datasets that are re-packaged, both the original license and the license of the derived asset (if it has changed) should be provided.
        \item If this information is not available online, the authors are encouraged to reach out to the asset's creators.
    \end{itemize}

\item {\bf New Assets}
    \item[] Question: Are new assets introduced in the paper well documented and is the documentation provided alongside the assets?
    \item[] Answer: \answerYes{} 
    \item[] Justification: New assets introduced in the paper consist of our codebase which includes notebooks to replicate our experiments and analyses, and contains documentation.
    \item[] Guidelines:
    \begin{itemize}
        \item The answer NA means that the paper does not release new assets.
        \item Researchers should communicate the details of the dataset/code/model as part of their submissions via structured templates. This includes details about training, license, limitations, etc. 
        \item The paper should discuss whether and how consent was obtained from people whose asset is used.
        \item At submission time, remember to anonymize your assets (if applicable). You can either create an anonymized URL or include an anonymized zip file.
    \end{itemize}

\item {\bf Crowdsourcing and Research with Human Subjects}
    \item[] Question: For crowdsourcing experiments and research with human subjects, does the paper include the full text of instructions given to participants and screenshots, if applicable, as well as details about compensation (if any)? 
    \item[] Answer: \answerNA{} 
    \item[] Justification: Our paper does not involve crowdsourcing nor research with human subjects.
    \item[] Guidelines:
    \begin{itemize}
        \item The answer NA means that the paper does not involve crowdsourcing nor research with human subjects.
        \item Including this information in the supplemental material is fine, but if the main contribution of the paper involves human subjects, then as much detail as possible should be included in the main paper. 
        \item According to the NeurIPS Code of Ethics, workers involved in data collection, curation, or other labor should be paid at least the minimum wage in the country of the data collector. 
    \end{itemize}

\item {\bf Institutional Review Board (IRB) Approvals or Equivalent for Research with Human Subjects}
    \item[] Question: Does the paper describe potential risks incurred by study participants, whether such risks were disclosed to the subjects, and whether Institutional Review Board (IRB) approvals (or an equivalent approval/review based on the requirements of your country or institution) were obtained?
    \item[] Answer: \answerNA{} 
    \item[] Justification: Our paper does not involve crowdsourcing nor research with human subjects.
    \item[] Guidelines:
    \begin{itemize}
        \item The answer NA means that the paper does not involve crowdsourcing nor research with human subjects.
        \item Depending on the country in which research is conducted, IRB approval (or equivalent) may be required for any human subjects research. If you obtained IRB approval, you should clearly state this in the paper. 
        \item We recognize that the procedures for this may vary significantly between institutions and locations, and we expect authors to adhere to the NeurIPS Code of Ethics and the guidelines for their institution. 
        \item For initial submissions, do not include any information that would break anonymity (if applicable), such as the institution conducting the review.
    \end{itemize}

\end{enumerate}

\end{document}